\pdfoutput=1

\documentclass[11pt]{article}

\usepackage{EMNLP2023}

\usepackage{times}
\usepackage{latexsym}

\usepackage[T1]{fontenc}

\usepackage[utf8]{inputenc}

\usepackage{microtype}

\usepackage{inconsolata}

\usepackage{comment}
\usepackage{times}
\usepackage{latexsym}
\usepackage{graphicx}
\usepackage{multirow}
\usepackage{subfig}
\usepackage{float}
\usepackage{amssymb}
\usepackage{nccmath}
\usepackage{textcomp}

\usepackage{amsmath}
\usepackage{xcolor}
\usepackage{pgfplots}
\usepackage{tikz}
\usetikzlibrary{shapes,arrows}
\usetikzlibrary{positioning}
\usepackage{wrapfig,lipsum,booktabs}

\usepackage{enumitem}
\usepackage{soul,color}
\usepackage{array}
\usepackage[ruled, lined, linesnumbered, commentsnumbered, longend]{algorithm2e}
\usetikzlibrary{patterns}

\usepackage{rotating}

\usepackage{pifont}
\newcommand{\cmark}{\ding{51}}
\newcommand{\xmark}{\ding{55}}%
\usepackage[font=normalsize,belowskip=-10pt,aboveskip=0pt]{caption}

\title{Adapt and Decompose: Efficient Generalization of Text-to-SQL via Domain Adapted Least-To-Most Prompting}

\author{Aseem Arora, Shabbirhussain Bhaisaheb, Harshit Nigam, \\ {\bf Manasi Patwardhan, Lovekesh Vig, Gautam Shroff}\\
        TCS Research, India \\
    \texttt{\{aseem.arora, shabbirhussain.b, h.nigam,}\\ \texttt{ manasi.patwardhan,  lovekesh.vig, gautam.shroff\}}@tcs.com}

\begin{document}
\maketitle
\begin{abstract}
Cross-domain and cross-compositional generalization of Text-to-SQL semantic parsing is a challenging task. Existing Large Language Model (LLM) based solutions rely on inference-time retrieval of few-shot exemplars from the training set to synthesize a run-time prompt for each Natural Language (NL) test query. In contrast, we devise an algorithm which performs offline sampling of a minimal set-of few-shots from the training data, with complete coverage of SQL clauses, operators and functions, and maximal domain coverage within the allowed token length. This allows for synthesis of a fixed Generic Prompt (GP), with a diverse set-of exemplars common across NL test queries, avoiding expensive test time exemplar retrieval. We further auto-adapt the GP to the target database domain (DA-GP), to better handle cross-domain generalization; followed by a decomposed Least-To-Most-Prompting (LTMP-DA-GP) to handle cross-compositional generalization. The synthesis of LTMP-DA-GP is an offline task, to be performed one-time per new database with minimal human intervention. Our approach demonstrates superior performance on the KaggleDBQA dataset,  designed to evaluate generalizability for the Text-to-SQL task. We further showcase consistent performance improvement of LTMP-DA-GP over GP, across LLMs and databases of KaggleDBQA, highlighting the efficacy and model agnostic benefits of our prompt based adapt and decompose approach.  
\end{abstract}


\section{Introduction}
Recently, Large Language Models (LLMs) such as GPT3\citep{brown2020language}, Codex\citep{Chen2021EvaluatingLL}, PaLM\cite{chowdhery2022palm}, pretrained with massive volumes of data have shown improved performance for multiple reasoning tasks using in-context learning \citep{Brown2020LanguageMA,Huang2022TowardsRI}, including program synthesis \citep{Austin2021ProgramSW,Jain2021JigsawLL,Nijkamp2022CodeGenAO} and semantic parsing \citep{Shin2021FewShotSP,Drozdov2022CompositionalSP,Shin2021FewShotSP,Shin2021ConstrainedLM}. There are a few recent approaches where LLMs are specifically used for Text-to-SQL semantic parsing in a (i) zero-shot setting \citep{Rajkumar2022EvaluatingTT,Chang2023HowTP,Nan2023EnhancingFT} where only the test Natural Language (NL) query constitutes the prompt, (ii) few-shot setting where exemplars similar to the test query in the target domain are retrieved from the available training data and appended to the test NL query to constitute the prompt \citep{Poesia2022SynchromeshRC,Chang2023HowTP,Nan2023EnhancingFT,An2023SkillBasedFS}. 
For this setting the available NL-SQL pairs would belong to domains that are different from the target database domain (iii) few-shot setting where exemplars are sampled NL-SQL queries available for the target-domain with maximum coverage of compositions \citep{Rajkumar2022EvaluatingTT,Qiu2022EvaluatingTI,Hosseini2022OnTC,yang-etal-2022-seqzero,Chang2023HowTP}. 
In this paper, we are mainly interested in (ii)  i.e. cross-domain generalization along with the scenario where the test queries may not have the set-of compositions covered in the training data (cross-composition generalizability). Moreover, considering a purely cross-domain setting, as opposed to (iii) above, we assume NO availability of exemplars belonging to the target databases. One solution to this setting is manual synthesis of few-shots for every new target database from scratch. However, this process is very tedious, time-consuming and also does not ensure diversity in the few-shots, with good coverage of SQL operators. Thus, there is a need for an efficient approach which can exploit available NL-SQL pairs from distinct domains, to intelligently sample few-shots and design prompts.

Synchromesh \citep{Poesia2022SynchromeshRC} and \cite{Nan2023EnhancingFT,An2023SkillBasedFS} have a similar setting, except they retrieve exemplars during run-time (during inference) by selecting NL queries as exemplars with similarity based on (a) NL query semantics \citep{Nan2023EnhancingFT}  or  (b) target SQLs (Target Similarity Tuning) \cite{Poesia2022SynchromeshRC} or (c) LLM generated `skill' based NL representation, focusing on program compositions and ignoring the surface NL forms \cite{An2023SkillBasedFS}. This reliance on inference-time retrieval of similar few-shots from the available data to build a run-time prompt and generate SQL for a test NL query, results in a less efficient solution. As opposed to this, we devise an algorithm which samples a minimal set-of few-shots from the training data ensuring complete coverage of SQL clauses, operators and functions and maximum coverage of database domains, which fits into the token length restriction. We append these few-shots with the out-of-distribution test NL query to define what we term as a \textit{Generic Prompt (GP)}, which is further used to generate the corresponding SQL. The \textit{GP} is generated offline and is common across distinct test queries, resulting in a more time-efficient solution obviating the need for real time retrieval. 
We further auto-adapt the GP to (a) the target database domain, and refer to it as domain adapted GP (\textit{DA-GP}), to better handle cross-domain generalization, (b) decompose it into a Least-To-Most-Prompting approach \cite{Zhou2022LeasttoMostPE} (\textit{LTMP-GP}) to better handle cross-compositional generalization, and (c) combine the approaches (a) and (b) to exploit their complementary benefits (\textit{LTMP-DA-GP}). In line with our motivation, formation of \textit{LTMP-DA-GP} is an efficient solution, as it is an offline task to be performed one-time per new database and is mostly programmatic with minimal human intervention (only needed for validation of prompts). We further demonstrate a consistent performance improvement of (a) and (b) over the base \textit{GP} and (c) over (a) and (b) on the databases of Kaggle-DBQA dataset \cite{lee-etal-2021-kaggledbqa}, designed to evaluate the generalizability of the Text-to-SQL task using distinct LLMs. Moreover, our approach not only yields an efficient solution, but also yields best performance for the used LLMs on KaggleDBQA. Following are our main contributions: 
\vspace*{-2mm} 
\begin{itemize}
\itemsep0em
    \item To the best of our knowledge, apart from \cite{Nan2023EnhancingFT} ours is the only approach to implement offline programmatic prompt generation ensuring diversity of samples for NL-to-SQL task. Moreover, as opposed to \cite{Nan2023EnhancingFT},  our \textit{GP} based sampling technique guarantees complete coverage of SQL operators and maximum converge of database domains. 
    \item Ours is the first approach of programmatic domain-adaptation of prompt, consistently showcasing performance improvement across multiple Kaggle-DBQA databases validating NL-to-SQL domain generalization capability.
    \item Ours is the first approach applying Least-to-Most-Prompting \cite{Zhou2022LeasttoMostPE}  for compositional generalization of complex NL-to-SQL task, showcasing consistent performance improvement across LLMs. 
    \item As compared to existing similarity based exemplar sampling approaches \citep{poesia2022synchromesh,Chang2023HowTP, An2023SkillBasedFS}, our approach of offline prompts synthesis proves to be more efficient.
    \item Our pipeline yields the best performance on the \textit{KaggleDBQA} dataset, reported in the literature for the used LLMs.
\end{itemize}

\begin{figure*}[t]
\begin{center}
\includegraphics[width=\textwidth]{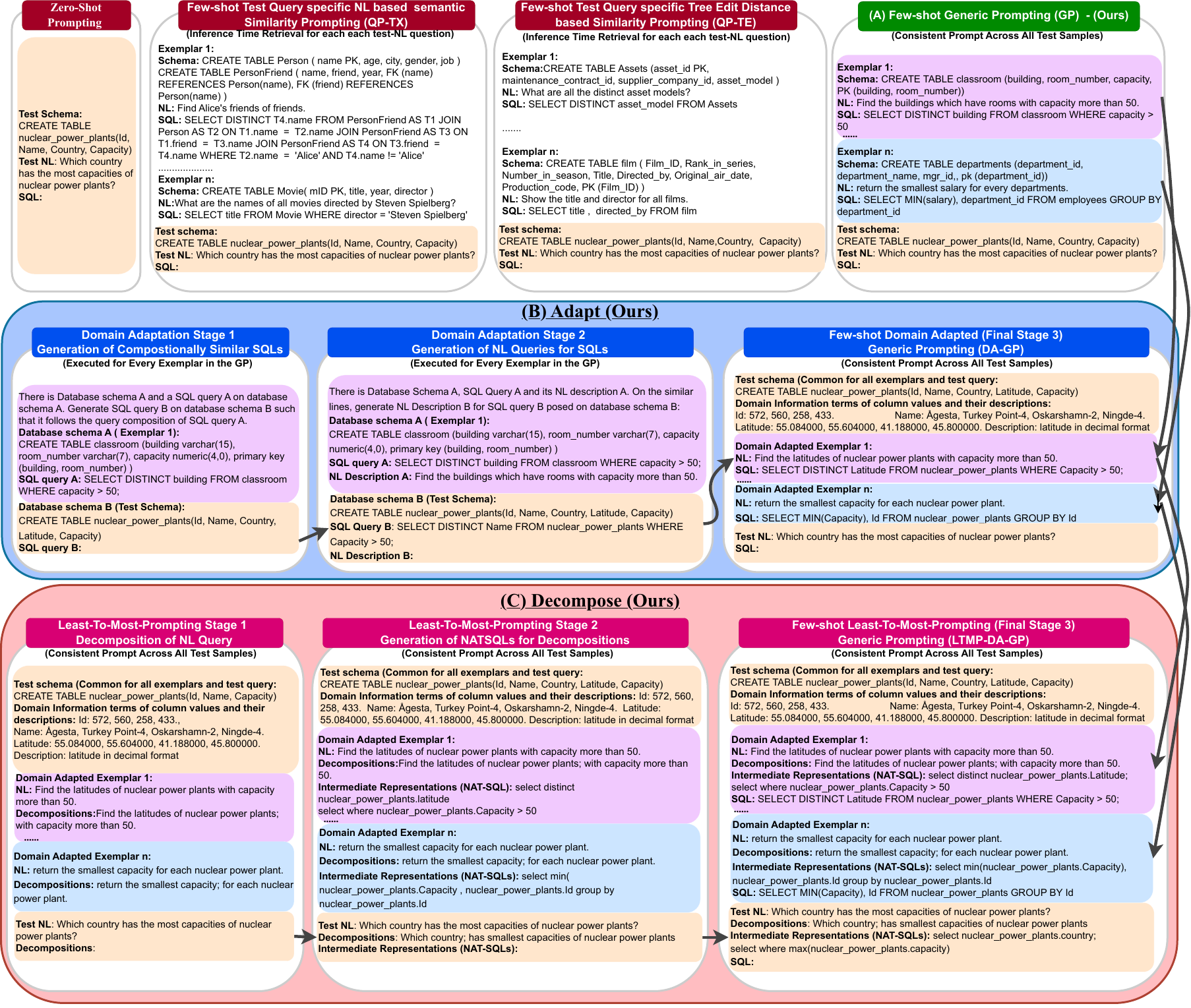}     \caption{Comparison of our Approach (A) GP (B) DA-GP  (Stage 1, 2, 3) and (C) LTMP-DA-GP (Stage 1, 2, 3) with Prior Zero-shot, Query-Similarity based Few-shot Approaches}
\label{fig:app}
\end{center} 
\end{figure*}

\section{Datasets}\label{sec:ds}


\textbf{Spider} \cite{gan2022measuring}: Is a large-scale, complex, and cross-domain Text-to-SQL benchmark dataset with a total of 200 databases (140, 20, 40 in the training, development and test splits with 7000, 1034, and 2147 Text-to-SQL pairs). We use state-of-the-art models trained on Spider to benchmark performance of our approach against supervised approaches. We also use the training split of \textit{Spider} to sample the exemplars, which serve as few-shots in our prompt (Section \ref{sec:app}).

\textbf{Spider-CG} \cite{gan2022measuring}: This dataset is designed for evaluating Text-to-SQL compositional generalization performance. To synthesize this dataset, Text-to-SQL pairs from \textit{Spider-Train} are transformed to corresponding sub-sentences and NatSQL pairs to form \textit{Spider-SS} dataset. The sub-sentences are obtained using a sentence-split algorithm and the corresponding NatSQL, which is an intermediate representation of SQL, is manually annotated. 
We use the \textit{Spider-SS} for the Least-To-Most-Prompting (LTMP) experiment, by retrieving the NL query decompositions in terms of sub-sentences and the corresponding intermediate NatSQL representations for few-shots sampled from  \textit{Spider-Train}, to synthesize prompts for each stage of LTMP (Section \ref{sec:LTMPGP}).

\textbf{Kaggle-DBQA} \cite{lee-2021-kaggle-dbqa}: This is a cross-domain  Text-to-SQL evaluation dataset with real world databases (DB). It covers a total of 8 DBs, viz. (i) Nuclear (10,22), (ii) Crime (9,18), (iii) Pesticide (16, 34), (iv) MathScore (9,19), (v) Baseball (12,27), (vi) Fires (12, 25), (vii) WhatCD  (13,28), and (viii) Soccer (6,12), indicating fine-tuning and test examples, respectively.  As we assume a purely cross-domain setting with no availability of in-domain few-shots, we do not use the fine-tuning samples but use only the test samples to evaluate our proposed LLM based pipelines. We prefer Kaggle-DBQA over other cross-domain and composition evaluation datasets such as spider-CG \cite{gan2022measuring}, because it is a real-web dataset and contains fewer test queries allowing us to showcase the efficacy of our approach using commercial LLMs with low cost overheads.

\section{Large Language Models (LLMs)} \label{sec:llm}
The literature has demonstrated state-of-the-art few-shot performance for NL-to-SQL \cite{Rajkumar2022EvaluatingTT,Nan2023EnhancingFT,Chang2023HowTP}  tasks, with Codex \cite{https://doi.org/10.48550/arxiv.2107.03374} (Code-da-Vincci) \cite{Chen2021EvaluatingLL} and GPT4\footnote{https://platform.openai.com/docs/models/gpt-4} from OpenAI. However, Codex is discontinued by OpenAI \footnote{Starting 23rd March, 2023.} and the cost of GPT4 is very high. Instead, we use two LLMs to demonstrate the efficacy of the proposed pipeline, viz. OpenAI`s  GPT-3.5-Turbo (ChatGPT), which is 60x cheaper than GPT4 and Text-da-Vincci-003, which is 6x cheaper than GPT4, both with 175B  parameters and 4096 token length restriction \footnote{https://platform.openai.com/docs/models/gpt-3-5}. 
We have not used other open-source models pretrained on code, such as CodeT5+ \cite{Wang2023CodeT5OC} (768 token length) or Codegen (2048 token length) \cite{Nijkamp2022CodeGenAO} for our study as they do not offer the token length required for our prompt (4K). 

\section{Approach}\label{sec:app}

\subsection{Database Schema Format} \label{sec:DBformat}
As the part of the prompt design, an important choice we make is the format of the schema of the Databases(DBs) to which the sampled few-shot exemplars and the test query belong. Each of our experiment, we keep the format of the DBs for the few-shots and the test (target) DB to be consistent. Existing approaches, yield state-of-the-art zero and few-shot results on \textit{Spider-Dev} dataset and its variants  using \textit{CREATE TABLE} schema format with (i) TEN selected rows \citep{rajkumar2022evaluating}, (ii) THREE column values \citep{Chang2023HowTP} or (iii) semantic augmentation with blocked column descriptions \citep{Nan2023EnhancingFT}. These approaches can afford to use these elaborate schema formats because of the use of Codex and GPT4 LLMs with allowable $>8K$ token lengths. As opposed to this, we use models with smaller allowable token length of maximum $4K$ (Section Section \ref{sec:llm}). To allow the the inclusion of few-shot exemplars in the \textit{Generic Prompt} sampled by our algorithm (Section \ref{sec:GP}) along with the reasoning  Least-to-Most-Prompting (LTMP) stages (\textit{LTMP-GP}: Section \ref{sec:LTMPGP}), we compromise on the elaborate schema format to fit the prompt in the allowable token length. For the \textit{GP} and \textit{LTMP-GP} based pipelines we use the \textit{CREATE TABLE} database schema format with Primary-Key and Foreign-Key constraints, but without the mention of column data-types and inclusion of row/column values and descriptions (Figure \ref{fig:app}). 
With domain adaptation of the \textit{GP} to the target database (Section \ref{sec:DAGP}), we require inclusion of only one (target) schema in the prompt allowing us to use an elaborate schema format. Thus, for the domain adaptation \textit{DA-GP} and it's further enhancement with LTMP (LTMP-DA-GP: Section \ref{sec:LTMPDAGP}), we include the domain information to the \textit{CREATE TABLE} format (Figure \ref{fig:app}) in the form of (i) column data-types, (ii) randomly sampled FOUR values for categorical and date-time columns, (iii) range of values for numerical columns and (iv) additional column descriptions, wherever necessary. 

\subsection{Generic Prompt (GP) Design Algorithm}
\label{sec:GP}
We have defined Algorithm \ref{algo} to sample the few-shots to form Generic Prompt (\textit{GP}). The algorithm is designed to select exemplars from a dataset with available text-SQL annotations, to ensure complete coverage of SQL clauses, operators and functions and maximum coverage of domains (databases) which can fit into the allowable token length. 
We assume to have an annotated dataset $D = \{db_j,\{t_{ij},s_{ij},a_{ij}\}_{i=1}^{N_j}\}_{j=1}^{M}$, where $t$ and $s$ are the annotated text-SQL query pairs posed on databases $db$ and $a$ are the answers of the SQL queries after execution, $N_j$ are the query pairs of database $db_j$, $M$ are total number of databases.
We have a test dataset  
 $T = \{db_l,\{t_{kl},s_{kl},a_{kl}\}_{k=1}^{K_l}\}_{l=1}^{L}$ of 
$\sum^L_{l=1}K_l$ query pairs and $L$ databases, such that 
$\{db_j\}_{j=1}^M \cap \{db_l\}_{l=1}^L = \phi$. Thus, as we consider completely cross-domain setting, we do not have any overlap between the training and test databases. We manually collect SQL operators, clauses and functions covered by queries  $s_{ij} \in D$ to form a set-of primitive operations $O$, including (i) SQL Clauses (`FROM', `HAVING', `WHERE', `ORDER BY', etc),  (ii) SQL Operators such as arithmetic (+, -, *, /, \%), comparison (=, !=, $<$, $>$, etc) and logical (ALL, AND, ANY, LIKE, etc) and (iii) SQL Functions (AVG, COUNT, MAX, MIN, etc). 

To sample the few-shot exemplars $E$ for the \textit{GP}, we sort the databases  ${db_j} \in D$ based on the operator coverage by the SQL queries ${s_{ij}}$. 
We perform query-pair (sample) traversal of this ordered list of databases. A sample $\{db_j,t_{ij},s_{ij}\}$ becomes part of $E$, if $s_{ij}$ covers at the least one uncovered primitive operation in $O$. If $s_{ij}$ covers a super-set of primitive operations of any query $s_x$ in an existing exemplar $\{db_x, t_x,s_x\} \in E$ then this exemplar is replaced by $\{db_i,t_{ij},s_{ij}\}$. The algorithm terminates when all the possible primitive operations in $O$ are covered by exemplars in $E$. This algorithm allows us to sample a minimal set of query-pairs as exemplars covering the complete set of primitive operations. This brings in diversity in the compositions of the SQL queries chosen as exemplars. The database ordering as per the primitive operator coverage, ensures minimal set of DB schema to be added in the GP, exhausting less number of tokens. However, selection of multiple DBs achieves diversity in the domains covered. We append the sampled few-shot exemplars with a NL test query, which is a sample $\{db_l,t_{lk}\}\in T$, retaining the consistency in the schema representation, forming the \textit{GP} (Figure \ref{fig:app}).

\vspace*{-2mm} 

\begin{algorithm}[t]
\footnotesize 
\caption{Generic Prompt Creation}
\label{algo}
\SetKwInOut{KwIn}{Input}
\SetKwInOut{KwOut}{Output}
\SetKwInOut{KwIS}{Initial Stage}
\KwIn{
$D = \{db_j,\{t_{ij},s_{ij},a_{ij}\}_{i=1}^{N_j}\}_{j=1}^{M}$ 
\tcp{ Dataset with databases, text, SQL queries and answer tuples}
$T = \{db_l,\{t_{kl},s_{kl},a_{kl}\}_{k=1}^{K_l}\}_{l=1}^{L}$ \tcp{Test Set}
$O = \{Operators, Clauses, Functions\}$ 
  \tcp{Set of SQL Primitive Operators}
}
\KwOut{GP \tcp{Generic Prompt}}
\KwIS{ 

 $E \gets \Phi$ \tcp{Set of Exemplars}
}

    $D \gets $\textit{Sort}(\textit{Extract\_Operators}($db_j in D$))
    \tcp{Sort databases in D on operator coverage}
    \For{$s_{ij} \in D$}
    {
    $O_e \gets$ \textit{Extract\_Operators}($s_{ij}$) 
    \tcp{Extract operators from training SQL} 
    
        \If{$(o \in O_e) \in O$} 
        { 
            \tcp{an operator is not covered}

            \For{$s_x \in E$} 
                {
                    \If{\textit{Extract\_Operators}($s_x$) $ \subseteq  O_e$ 
                    }  
                    {
                        $E \gets E - \{db_x,t_x,s_x\}$ 
                        \tcp{ Remove existing exemplar with operators to be subset of current exemplar operators}
                    }
                }
                
            $E \gets E + \{db_j,t_{ij},s_{ij}\}$ 
            \tcp{Add the tuple as an Exemplar}
            $O \gets O - O_e$  \tcp{Remove covered operators}

        }

    }

$GP \gets E + \{db_l,t_{kl}\} \in T$

\end{algorithm}

\subsection{Domain Adaptation of GP (DA-GP)} 
\label{sec:DAGP}
As ours is a completely cross-domain setting, the \textit{GP} consists of domains defined by database for which the few-shots are sampled from the train-dataset, which are distinct from the target database. We auto-adapt these few-shots to the domain of the target database, keeping the query compositions consistent. The hypothesis is that the adaptation should facilitate the LLMs to achieve better performance on the queries of the target domain. This task is performed in three stages.

\textbf{Stage1: Generating Compositionally similar SQLs in the target domain:}
For each few-shot SQL query in the \textit{GP} we feed the serialized source schema and SQL (without NL) along with the serialized target schema and prompt the LLM to generate SQL on the target schema which is compositionally similar to the source query, by explaining what is compositional similarity. We sample SQL queries from the beam, until we find an executable SQL query on the target DB, whose skeleton has the tree edit distance to be within a threshold to that of the skeleton of the original few-shot SQL query. This ensures compositional similarity (Figure \ref{fig:app}).

\textbf{Stage2: Generating text queries for the SQL queries:}
We feed each compositionally similar SQL, generated for each few-shot exemplar in the \textit{GP}, to the LLM along with the target schema and prompts it to generate the NL question which describes the SQL query in text form.  This step allows us to have a NL-SQL pair in the target domain (database) for each few-shot exemplar in the \textit{GP}. 

\textbf{Stage 3: Using Domain Adapted GP to generate SQLs for the test NL queries:}
We form \textit{Domain Adapted Generic Prompt}(\textit{DA-GP}) using the target schema with the available domain information  (Section \ref{sec:DBformat}) and the domain specific NL-SQL few-shot pairs. Note that \textit{DA-GP} has one database schema consistent across the few-shots as well as the test query. We append the test NL query to \textit{DA-GP} and feed it to the LLM to generate SQL.

\begin{table*}[t]
 \centering
    \resizebox{\textwidth}{!}{
     \begin{tabular}{|l|l|c|c|c|c|c|c|c|c|c|}
     \hline
     \multirow{2}{*}{\textbf{Model} }& \multirow{2}{*}{\textbf{Approach}} & \multicolumn{8}{c|}{ \textbf{Database-wise Execution Accuracy (\%Ex)} }  & \textbf{Total} \\
      \cline{3-10}
     &          &  \begin{tabular}{c} Geo \\Nuclear \end{tabular}& \begin{tabular}{c} Greater \\ Manchester\end{tabular} & Pesticide & \begin{tabular}{c} Student \\Maths Score \end{tabular} & \begin{tabular}{c} The History \\of Baseball\end{tabular} & \begin{tabular}{c}  US \\Wildfires\end{tabular} & \begin{tabular}{c} What CD\\ Hiphop\end{tabular} & \begin{tabular}{c} World \\Soccer \end{tabular}& \% Ex\\
     \hline

      \hline
           \multirow{4}{*}{\begin{tabular}{l}GPT-Turbo-3.5 \\ (Benchmark) \end{tabular}}  & Zero-Shot   & 27.27  & 33.33 & 19.35 & 10.93 & 11.11 & 32.00 & 7.69 & 41.67 & 21.11\\
           ~ &   QP-TX $\dagger$    &  31.82 & 44.44 & 29.03 & 15.79 & 11.11 & 28.00 & 34.62 & 8.33 & 26.11 \\
          
          ~ &   QP-TE  $\dagger$  &   27.27 & 27.78 & 22.58 & 15.79& 25.93& 44.00 & 26.82& 33.33 & 27.78\\
           
           ~ &  DIN-SQL $\dagger$  &  - & -&  -& - & - & -& - &- & 27.00\\
           \hline
     \multirow{5}{*}{\begin{tabular}{l}Text-da-vincci-003 \\ (Benchmark) \end{tabular}} &   Zero-Shot    &   27.27 & 22.22 & 29.03 & 15.79& 18.52& 20.00 & 7.69& 8.33 & 19.44\\  
     ~ &  QP-TX $\dagger$  & 27.27  & 33.33 & 19.35 & 10.93 & 11.11 & 32.00 & 7.69 & 41.67 & 21.11\\
     ~ & QP-TE $\dagger$ & 27.27  & 33.33 & 19.35 & 10.93 & 11.11 & 32.00 & 7.69 & 41.67 & 21.11\\
      ~ &   DIN-SQL $\dagger$   &  45.45 & 33.33 & 20.59 & 21.05 & 18.52 & 50.00 & 21.43 & 25.00& 29.18\\
       ~ & QP-SK & -  & - &  - & - & - & - &  -& - & \textbf{36.80}\\
     \hline
     \hline
       \multirow{4}{*}{\begin{tabular}{l}GPT-Turbo-3.5 \\ \textbf{(Ours)}\end{tabular}}  & GP  &  31.82 &27.78  & 29.03  & 5.26  & 14.81 & 36.00 & 23.08 & 25.00 & 24.44 \\
           ~ &   DA-GP    & 40.91  &33.33 & \textbf{ 35.48}  & 10.53 & 22.22 &20.00 & 15.38 & 25.00  & 25.56 \\
           ~ &   LTMP-GP    &  27.27  & \underline{\textbf{50.00}} &  29.03 & 10.53 & 14.81 & 52.00 & 30.77 & 33.33  & 30.56 \\
     ~ &   LTMP-DA-GP   &  \textbf{50.00} & 22.22 & 32.26  &  \textbf{15.79} & \underline{\textbf{22.22}} & \underline{\textbf{60.00}} & \underline{\textbf{30.77}} & \underline{\textbf{33.33}} & \textbf{33.89} \\
           \hline
     \multirow{4}{*}{\begin{tabular}{l}Text-da-Vincci-003 \\\textbf{(Ours)} \end{tabular}}  & GP  &  31.82 & 22.22  & 35.48  & 10.53  & 11.11 & 20.00 & 15.38 & 16.67  & 21.11 \\
           ~ &   DA-GP    &  40.91 &27.78 & \underline{\textbf{45.16}}  & 21.05 & 18.52 & 44.00 & 15.38 &  25.00 & 30.56 \\
           ~ &   LTMP-GP    &  59.09  & 44.44 & 41.18 & 21.05 & 22.22 & 52.00  & 21.43  & 25.00  & 36.41\\
      ~ &   LTMP-DA-GP   & \underline{\textbf{63.64}}  & \textbf{44.44} & 41.18 &  \underline{\textbf{26.32}} & \underline{\textbf{22.22}} & \textbf{56.00} & \textbf{21.43} & \textbf{25.00}  & \underline{\textbf{38.04} }\\
           \hline

     \end{tabular}
    }
    \caption{Kaggle DBQA - Results. Comparison with zero and few-shot approaches, QP-TX: Query specific TeXt-based Similarity\cite{poesia2022synchromesh}, QP-TE:Query specific Tree-Edit-distance-based Similarity \cite{poesia2022synchromesh}, QP-SK: Query specific Skill based Similarity \cite{An2023SkillBasedFS}, DIN-SQL \cite{Pourreza2023DINSQLDI}, 
    $\dagger$Few-shots: 16, Result: \underline{\textbf{Overall Best}},\textbf{Best for the LLM} }
    \label{tab:results}
\end{table*}

\begin{table}[t]
 \centering
    \resizebox{\columnwidth}{!}{%
     \begin{tabular}{|l|l|c|}
     \hline
     \textbf{Model} & \textbf{Approach} & \textbf{\%EX} \\
  
     \hline
     
      RATSQL \cite{gan2022measuring}     &   \multirow{7}{*}{\begin{tabular}{l}Supervised\\ Trained with \\\textit{Spider-Train} \\\cite{gan2022measuring}\end{tabular}}       &  13.56\\
      T53B \cite{Lan2023UNITEAU}     &     ~     &   26.80 \\
      SmBOP \cite{rubin2020smbop}     &     ~     &   27.20\\
      RASAT \cite{Qi2022RASATIR}     &     ~     &    27.60 \\
      Picard \cite{scholak2021picard}     &     ~     & 29.80\\
       REDSQL \cite{Li2023RESDSQLDS}     &     ~     & 31.90\\
      UL-20B \cite{Lan2023UNITEAU}     &     ~      & 34.90 \\
           \hline
    RASAT \cite{rubin2020smbop}     &     \multirow{3}{*}{\begin{tabular}{l}Supervised\\ Trained on \textit{UNITE}\\\cite{Lan2023UNITEAU}\end{tabular}}    & 26.80\\
      T53B \cite{Lan2023UNITEAU}     &     ~      & 33.80 \\
      Picard \cite{scholak2021picard}     &     ~  & \textbf{36.80}\\
      \hline
          
           
           \hline
     \hline
     
       GPT-Turbo-3.5 \textbf{(Ours)} & LTMP-DA-GP  & 33.89\\
          
           
     Text-da-Vincci-003 \textbf{(Ours)}&  LTMP-DA-GP   & \underline{\textbf{38.04} }\\
           \hline

     \end{tabular}
    }
    \caption{Kaggle DBQA - Exec. Acc. \underline{\textbf{Best}}, \textbf{Second Best}. Supervised 
    Approaches Comparision }
    \label{tab:supresults}
\end{table}

\subsection{Least-to-Most-Prompting with GP (LTMP-GP)} 
\label{sec:LTMPGP}
The \textit{GP} covers all the primitive SQL Operations, Clauses and Functions. However, few-shots cover only a few compositions of these primitive operations. We perform LTMP to help the LLMs achieve generalizability on compositions unseen in the few-shots as well as in the pre-training data. To achieve this, we decompose the NL-to-SQL task into the following three sub-tasks and semi-auto-adapt each of the few-shot exemplars in the \textit{GP} for each of the following sub-tasks (Figure \ref{fig:app}). The hypothesis is that the decomposition of few-shots, helps exposing the underlying NL-SQL mappings at more primitive level, through the NAT-SQL based intermediate representations, which can be reused by the LLMs to synthesize SQLs for unseen compositions.

\textbf{Stage 1: NL Query Decomposition:}
 The few-shot NL queries in \textit{GP} sampled from \textit{Spider-Train}, along with their decompositions fetched from the \textit{Spider-SS} (Section \ref{sec:ds}) forms the prompt for the first stage of LTMP.  We append it with the test NL query to generate the decomposition for the same. For exploiting the reasoning capabilities of LLMs, for each few-shot, we manually include the Chain-Of-Thoughts (COT) behind the decomposition of the NL queries, in terms of explaining the choice of split point (semantic segmentation) of the NL. 

\textbf{Stage 2: Mapping of NatSQL to NL decomposition:}
The few-shot NL queries in \textit{GP} sampled from \textit{Spider-Train}, along with their decompositions and NAT-SQLs, which is an intermediate representations of ground truth SQLs, fetched from  \textit{Spider-SS} (Section \ref{sec:ds}) forms the prompt for the second stage of LTMP. For each few-shot, we explain the COT behind the mapping of each decomposed NL query to the NatSQL, in terms of selection of the SQL clause for the NatSQL (part of the skeleton of NatSQL) and schema linking including specific table(s) and Column(s) selected for the NL decomposition to form the NatSQL. We append the prompt with the test NL query followed by its decompositions generated in the prior stage, to generate the NatSQL for each decomposition.

\textbf{Stage 3: Generating SQL from NatSQL:}
We auto-generate the third stage prompt of \textit{LTMP-GP} to include the few-shot NL queries in the \textit{GP} with their decompositions, corresponding NatSQLs and the ground truth SQLs. We append this with the test NL query, its decompositions and corresponding NatSQLs generated in the prior stages to generate the SQL for the test NL.


\subsection{LTMP with DA-GP (LTMP-DA-GP)} \label{sec:LTMPDAGP}
To exploit the complementary advantages of domain adaptation (domain generalization) and least-to-most prompting (compositional generalization), we perform LTMP over \textit{DA-GP} with executing all the three stages explained in the Section \ref{sec:LTMPGP} 
to construct the \textit{LTMP-DA-GP} 
. For this the domain adapted NL query decompositions and NAT-SQLs are manually created. We append the test NL query to the prompt of the first stage and the outputs of the prior stages to the subsequent stages recursively, as explained in Section \ref{sec:LTMPGP}, to finally generate the SQL as the result of the last stage.

\section{Results and Discussion}

\subsection{Benchmarks} \label{sec:bench}

\textbf{State-of-the-art supervised approaches}: Include models (Table \ref{tab:supresults} 
trained with \textit{Spider-Train} and \textit{UNITE} \cite{Lan2023UNITEAU} datasets and yielding SOTA results on \textit{Spider-Dev} and its variants. 

\textbf{Zero-shot approaches}:  \citep{rajkumar2022evaluating,Chang2023HowTP,Nan2023EnhancingFT}, yield SOTA results on \textit{Spider-Dev} and its variants with Codex and GPT4 as LLMs. For fair comparison, we compute zero-shot \textit{KaggleDBQA} results with our schema format and LLMs. 

\begin{table*}[!ht]
\centering
\resizebox{0.97\textwidth}{!}
{
\begin{tabular}{|p{0.75em}|p{1em}|p{1em}|p{2em}|p{1em}|p{4em}|p{60em}|p{2em}|}
\hline
&\textbf{GP} & \textbf{DA} & \small\textbf{LTMP -DA} & \textbf{No.} & \textbf{Category} & \textbf{Illustrative Samples} &  \textbf{\%} \\ \hline


\multirow{10}{0.75em}{\rotatebox{90}{\textbf{Case 1}}} & \multirow{10}{1em}{\xmark}  & \multirow{10}{1em}{\cmark} & \multirow{10}{1em}{\cmark}& 1 & Rectifying values & \textit{\textbf{NL}: Show all fires caused by campfires in Texas. \newline \textbf{GT}: SELECT * FROM Fires WHERE STAT\_CAUSE\_DESCR = "Campfire" AND State = "TX" \newline \textbf{GP}: SELECT  * FROM Fires WHERE STAT\_CAUSE\_DESCR = 'Campfire' AND STATE = \textcolor{red}{'Texas'}\newline \textbf{DA}:SELECT  * FROM Fires WHERE STAT\_CAUSE\_DESCR = 'Campfire' AND STATE = \textcolor{green} {'TX'}\newline \textbf{LTMP-DA}:SELECT * FROM Fires WHERE STAT\_CAUSE\_DESCR = "Campfire" AND STATE = \textcolor{green} {"TX"}
} &  5.38\\
\cline{5-8}
&  & &  & 2 & Rectifying columns & \textit{\textbf{NL}: How many number of units are there in sample 9628? \newline \textbf{GT}: SELECT quantity FROM sampledata15 WHERE sample\_pk = 9628 \newline \textbf{GP}: SELECT  \textcolor{red}{conunit} FROM resultsdata15 WHERE sample\_pk = 9628 \newline \textbf{DA-GP}:SELECT \textcolor{green} {quantity} FROM sampledata15 WHERE sample\_pk = 9628\newline \textbf{LTMP-DA}:SELECT \textcolor{green}{quantity} FROM sampledata15 WHERE sample\_pk = 9628
}  & 4.61\\
\hline

\multirow{15}{0.1em}{\rotatebox{90}{\textbf{Case 2}}} &  \multirow{15}{1em}{\xmark} & \multirow{15}{1em}{\xmark} & \multirow{15}{1em}{\cmark} & 1 & Extra SELECT + Operation on Column & \textit{\textbf{NL}: Which country lead the total capacity of the power plants it held? \newline \textbf{GT}: SELECT Country FROM nuclear\_power\_plants GROUP BY Country ORDER BY sum(Capacity) DESC LIMIT 1 \newline \textbf{GP}: SELECT  \textcolor{red}{Country, SUM(Capacity)} AS TotalCapacity FROM nuclear\_power\_plants GROUP BY Country ORDER BY TotalCapacity DESC LIMIT 1\newline \textbf{DA}: SELECT  \textcolor{red}{Country, SUM(Capacity)} AS TotalCapacity FROM nuclear\_power\_plants GROUP BY Country ORDER BY TotalCapacity DESC LIMIT 1 \newline \textbf{LTMP-DA}: SELECT \textcolor{green}{Country} FROM nuclear\_power\_plants GROUP BY Country ORDER BY SUM(Capacity) DESC LIMIT 1
}  & 5.51\\
\cline{5-8} & &  & 
 & 2 & Correcting Operations & \textit{\textbf{NL}: Which country has the least capacities of nuclear power plants? \newline \textbf{GT}: SELECT Country FROM nuclear\_power\_plants GROUP BY Country ORDER BY sum(Capacity) LIMIT 1 \newline \textbf{GP}: SELECT  Country, \textcolor{red}{MIN(Capacity)} FROM nuclear\_power\_plants GROUP BY Country ORDER BY Capacity ASC LIMIT 1 \newline \textbf{DA}: SELECT  Country, \textcolor{red}{MIN(Capacity)} FROM nuclear\_power\_plants GROUP BY Country ORDER BY Capacity ASC LIMIT 1 \newline \textbf{LTMP-DA}:SELECT Country FROM nuclear\_power\_plants GROUP BY Country ORDER BY \textcolor{green} {SUM(Capacity)} ASC LIMIT 1
}  & 0.69\\
\cline{5-8} & &  & 
 & 3 & Rectifying Values;Extra Condition & \textit{\textbf{NL}: In 2014, how many wildfires were the result of mismanaged campfires? \newline \textbf{GT}: SELECT count(*) FROM Fires WHERE STAT\_CAUSE\_DESCR LIKE "\%Campfire\%" AND FIRE\_YEAR = 2014 \newline \textbf{GP}:SELECT  COUNT(*) FROM Fires WHERE \textcolor{red} {STAT\_CAUSE\_DESCR = 'Miscellaneous' AND STAT\_CAUSE\_CODE = 13} AND FIRE\_YEAR = 2014\newline \textbf{DA}:SELECT  count(*) FROM Fires WHERE FIRE\_YEAR = 2014 AND \textcolor{red} {STAT\_CAUSE\_DESCR = 'Mismanaged Campfire'} \newline \textbf{LTMP-DA}: SELECT COUNT(*) FROM Fires WHERE FIRE\_YEAR = 2014 AND \textcolor{green}{STAT\_CAUSE\_DESCR = "Campfire"}}
 & 3.45\\
\hline

\multirow{15}{1em}{\rotatebox{90}{\textbf{Case 3}}} & \multirow{15}{1em}{\xmark} &\multirow{15}{1em}{\xmark} & \multirow{15}{1em}{\xmark} & 1 & Additional column & \textit{\textbf{NL}: What’s the most common type of crime? \newline \textbf{GT}: SELECT Type FROM GreaterManchesterCrime GROUP BY Type ORDER BY count(*) DESC LIMIT 1; \newline \textbf{GP}: SELECT Type, \textcolor{red}{COUNT(*)} AS Frequency FROM GreaterManchesterCrime GROUP BY Type ORDER BY Frequency DESC LIMIT 1 \newline \textbf{DA}: {SELECT  Type, \textcolor{red}{COUNT(*)} FROM GreaterManchesterCrime GROUP BY Type ORDER BY COUNT(*) DESC LIMIT 1 \newline \textbf{LTMP-DA}: SELECT Type, \textcolor{red}{COUNT(*)} FROM GreaterManchesterCrime GROUP BY Type ORDER BY COUNT(*) DESC LIMIT 1}
} & 42.26\\
\cline{5-8}  &  &  &  
& 2 & Logically incorrect & \textit{\textbf{NL}: How many matches in Spain in 2010? \newline \textbf{GT}: SELECT count(*) FROM football\_data WHERE \textcolor{green}{Season} LIKE "\%2010\%" AND Country = "Spain"; \newline \textbf{GP}: SELECT  COUNT(*) FROM betfront WHERE country = 'Spain' AND \textcolor{red}{YEAR} = 2010 \newline \textbf{DA}: SELECT  COUNT(*) FROM football\_data WHERE Country = 'Spain' AND \textcolor{red}{YEAR}(Datetime) = 2010 \newline \textbf{LTMP-DA}: SELECT COUNT(*) FROM football\_data WHERE Country = 'Spain' AND \textcolor{red}{YEAR} = '2010'
}& 11.22 \\
\cline{5-8} & & &
& 3 & Miscell-aneous & \textit{\textbf{NL}: What are the downloaded numbers and their release types? \newline \textbf{GT}: SELECT sum(totalSnatched), releaseType FROM torrents GROUP BY releaseType; \newline \textbf{GP}: SELECT \textcolor{red}{totalSnatched}, releaseType FROM torrents \newline \textbf{DA}: SELECT  \textcolor{red}{totalSnatched}, releaseType FROM torrents \newline \textbf{LTMP-DA}: SELECT \textcolor{red}{totalSnatched}, releaseType FROM torrents
}  & 22.06 \\
\hline

\multirow{6}{1em}{\rotatebox{90}{\textbf{Case 4}}} & \multirow{6}{1em}{\cmark} & \multirow{6}{1em}{\xmark} & \multirow{6}{1em}{\xmark} & 1 & Semantics Misinterpretation & \textit{\textbf{NL}: State with highest average math score \newline \textbf{GT}: SELECT state FROM NDECoreExcel\_Math\_Grade8 ORDER BY average\_scale\_score DESC LIMIT 1 \newline \textbf{GP}:SELECT \textcolor{green}{state} FROM NDECoreExcel\_Math\_Grade8 ORDER BY average\_scale\_score DESC LIMIT 1 \newline \textbf{DA}:SELECT  state, \textcolor{red}{MAX(average\_scale\_score)} FROM NDECoreExcel\_Math\_Grade8 GROUP BY state \newline \textbf{LTMP-DA}:SELECT state, \textcolor{red}{MAX(average\_scale\_score)} FROM NDECoreExcel\_Math\_Grade8 GROUP BY state ORDER BY MAX(average\_scale\_score) DESC LIMIT 1
}  & 1.37\\
\hline
\multirow{12}{1em}{\rotatebox{90}{\textbf{Case 5}}} & \multirow{12}{1em}{\cmark} & \multirow{12}{1em}{\cmark} & \multirow{12}{1em}{\xmark} & 1 & Fail to Understand Question; Incorrect Value & \textit{\textbf{NL}:What is the result in case 6B:E2:54:C6:58:D2? \newline \textbf{GT}:SELECT Outcome FROM GreaterManchesterCrime WHERE CrimeID = "6B:E2:54:C6:58:D2" \newline \textbf{GP}:SELECT \textcolor{green}{Outcome} FROM \textcolor{green}{GreaterManchesterCrime WHERE CrimeID = '6B:E2:54:C6:58:D2'} \newline \textbf{DA}:SELECT \textcolor{green}{Outcome} FROM \textcolor{green}{GreaterManchesterCrime WHERE CrimeID = '6B:E2:54:C6:58:D2'} \newline \textbf{LTMP-DA}:\textcolor{red}{SELECT * FROM Table}
}  & 2.76\\
 \cline{5-8} & &  & 
 & 2 & Syntax Error & \textit{\textbf{NL}: What's the code for confirmation for the latest sample? \newline \textbf{GT}: SELECT confmethod FROM resultsdata15 as T2 JOIN sampledata15 as T1 ON T1.sample\_pk = T2.sample\_pk ORDER BY year, month, day DESC LIMIT 1 \newline \textbf{GP}: SELECT  confmethod FROM resultsdata15 ORDER BY sample\_pk DESC LIMIT 1\newline \textbf{DA}: SELECT  confmethod FROM resultsdata15 ORDER BY sample\_pk DESC LIMIT 1 \newline \textbf{LTMP-DA}: SELECT confmethod FROM resultsdata15 WHERE sample\_pk = (SELECT sample\_pk FROM sampledata15 ORDER BY year \textcolor{red}{DESC}, month \textcolor{red}{DESC}, day DESC LIMIT 1)} & 0.69\\
\hline
\hline
\end{tabular}
}
\caption{Qualitative analysis. NL:Natural Language, GT: Ground Truth \cmark: \textcolor{green}{Correct SQL} and \xmark : \textcolor{red}{Incorrect SQL}. \% of Erroneous Test Queries}

\label{tab:qualitative analysis}
\end{table*}
    
\textbf{Existing few-shot approaches:}  Few-shots are sampled using Top-K samples from the train set using following sampling strategies found in the literature. For fair comparison, we choose the number few-shots (K) to be the same as the number of exemplars in the \textit{GP}.
(i) Test Query specific TeXt-based Similarity sampling (QP-TX) \cite{poesia2022synchromesh}: having maximum semantic similarity \cite{Reimers2019SentenceBERTSE} with the test NL query, (ii) Test Query specific Tree-Edit-distance-based Similarity sampling (QP-TE) \cite{poesia2022synchromesh}:  having maximum target program based similarity with the test NL query. Following Synchromesh \cite{poesia2022synchromesh}, we train Sentence BERT \cite{Reimers2019SentenceBERTSE}) as a scoring function to compute the tree-edit distance between the corresponding SQLs of the input NL queries (Target Semantic Tuning (TST)), (iii) Test Query specific Skill based similarity sampling (QP-SK): maximum skill based similarity \cite{An2023SkillBasedFS}. Here LLMs are used to retrieve skill based representations of the queries, by eliminating unimportant surface features. (iv) Diversity based sampling: \cite{Nan2023EnhancingFT} performs diversity sampling by picking up the exemplars near the centroids of the training sample clusters, formed using a combination of continuous NL embedding and discrete embedding with binary features representing syntactic elements of the SQL counterpart, including keywords, operators, and identifiers. Our \textit{GP} based approach not only selects diverse samples, but also ensures SQL operator coverage. We have not benchmarked against this approach due to unavailability of the prompts or the code. 
    

  \textbf{Chain-of-Thoughts (COT) approaches:} DIN-SQL \citep{Pourreza2023DINSQLDI} performs the NL-to-SQL task by dividing it into stages, viz. schema linking, NL query classification based on difficulty, distinct well-curated COTs prompting for distinct difficulty levels, along with few-shots with COT explanations and self-refinement at the end. We use the prompts in the paper for computing results. 

Note that for fair comparison, we have not benchmarked \textit{KaggleDBQA} results against the of LLMs not used the experimentation \cite{An2023SkillBasedFS,Chang2023HowTP,Nan2023EnhancingFT} .

\subsection{Experimentation and Results}
 We use the \textit{Spider-train} set as the training set to fetch few-shots for our \textit{GP} and \textit{Kaggle-DBQA} test set to evaluate the performance. Our algorithm yields 16 exemplars as few-shots covering a total of 4 databases. The selected queries cover a total of 32 SQL operators and clauses. For more deterministic results, we set the LLM parameters temperature to be 0. 
 
 The results are illustrated in the Table \ref{tab:results}. Our basic \textit{GP} performs better than (i) some state-of-the-art supervised models, (ii) zero-shot  and (iii) QP-TX. This is possible due to generalization capabilities of LLMs along with  programmatically sampled diverse few-shots in \text{GP}. Our final adapted and decomposed \textit{LTMP-DA-GP} consistently performs better than (i) All state-of-the-art supervised benchmarks and (ii) All few-shot benchmarks. This is due to the combined effect of diversity based sampling in GP \cite{Nan2023EnhancingFT,An2023SkillBasedFS} and effect of domain adaptation and LTMP. Except \textit{US Wildfires} database \textit{DA-GP}, consistently performs better than \textit{GP} showcasing the effect of domain adaptation for domain generalizability. \textit{LTMP-GP} consistently performs better than \textit{GP}, showcasing the effect of LTMP for compositional generalizability. Except \textit{Greater Manchester} and \textit{Pesticide} DBs \textit{LTMP-DA-GP} consistently improves over \textit{LTMP-GP} and \textit{DA-GP}, demonstrating complementary benefits of DA and LTMP for certain queries and thus proves the efficacy of our adapt and decompose pipeline. We find for \textit{Greater Manchester} DB the GPT-Turbo-3.5 performance drops with LTMP-DA-GP as with additional DA information the model tries to reason better generating long COTs, leading unavailability of tokens left to generate the desired output.

\subsection{Qualitative Analysis}

We manually analyze the test-queries (Table \ref{tab:qualitative analysis}). 
Case 1 Category (1) and (2) demonstrates samples where inclusion of domain knowledge in terms of table values and column descriptions rectifies the SQLs.
 Case 2 demonstrates LTMP rectifying samples due to better resolution of decomposed queries (e.g. 'Which country' and  'lead the total capacity of the power plants it held?') as opposed to the need of resolving complete query at once, with the prior approaches.
Case 3 are erroneous samples, where (1) LTMP can not fix additional aggregation operation appearing in the SELECT clause, especially where the NLs can not be decomposed or  generation of logically incorrect queries due to insufficient domain information such as: (2) query specific values (eg. 2010) not being present in the sampled values of the schema column descriptions (eg. season) in the prompt (wrong column `Year' gets picked up due to its specified range as 2009-2013) and (3) absence of understanding of domain specific numerical formula `downloaded numbers = sum(total snatched)' for songs for CD Hiphop DB. 
There are very few samples for Case 4 (1) where additional aggregation operation is added as a part of SELECT clause due to mis-interpretation of the NL query semantics. Eg. `state with' is been interpreted as providing some additional information along with `state' by LLM with DA as well as LTMP.
Case 5 (1) LTMP fails to decompose the question due to complex value further propagating error in the following stages (2) Queries come out correct for GP and DA due to samples being arranged in order of time, however with decompostions LTMP tried to come up with a right query but fails due to adding extra DESC condition to each column as following a few-shot decomposition.


\section{Conclusion}
In this paper, we leverage LLMs for the cross-domain and cross-composition generalization of Text-to-SQL. As opposed to prior approaches, which rely on inference-time retrieval of exemplars similar to the test query; we devise an algorithm which samples diverse set-of exemplars with complete coverage of SQL operators, clauses and functions and maximal coverage of databases to form the \textit{Generic Prompt (GP)}, which is common across every test sample obviating the need for dynamic exemplar retrieval and thus leading to an efficient approach.  We further perform programmatic domain-adaptation of this prompt \textit{DA-GP}, which consistently showcases performance improvement across multiple databases and LLMs better achieving domain generalization. We further decompose the exemplars of \textit{DA-GP}, to execute a novel pipeline of Least-to-Most-Prompting (\textit{LTMP-DA-GP}) for compositional generalization of the complex NL-to-SQL task. This pipeline showcases consistent improvement over \textit{GP} across multiple databases and LLMs demonstrating complementary benefits of the adapt and decompose steps and thus proving the efficacy of our approach.  Our pipeline, being offline with minimal human intervention, is not only efficient; but also yields the best performance reported in the literature  with the experimented LLMs, on KaggleDBQA dataset designed to test generalizability of NL-to-SQL task.




\bibliography{anthology,MainPaper}
\bibliographystyle{acl_natbib}

\appendix
\section{Supplementary Material}

\subsection{Existing Work on LLM based NL-to-SQL}\label{sec:rw}
\cite{Rajkumar2022EvaluatingTT,Chang2023HowTP,Nan2023EnhancingFT} has attempted to use LLMs for Text-to-SQL semantic parsing task in zero-shot as well as few-shot settings. For zero-shot setting, they experiment with various formats of the database schema, such as the APIDocs 
or SQL `\textit{CREATE TABLE}' commands, with and without randomly selected data rows or columns from the database table (elaborated in Section \ref{sec:DBformat}). For the few-shot setting, \cite{Rajkumar2022EvaluatingTT,Qiu2022EvaluatingTI,Hosseini2022OnTC,yang-etal-2022-seqzero} focus on cross-composition generalization and provide the queries which are posed on the target database itself as exemplars (cross-domain setting is not considered). Thus, the assumption is that few queries are available for a new database. They work with datasets such as GeoQuery (\cite{Tang2001UsingMC,Zelle1996LearningTP}) with data in US geography domain, Scholar (\cite{iyer-etal-2017-learning}) with data in academic publications or a dataset designed for queries in E-commerce domain (\cite{yang-etal-2022-seqzero}). In our approach, we assume no availability of annotated data in terms of SQL programs for the test database and thus, completely a cross-domain setting.

On the similar lines of our work, Synchromesh \citep{Poesia2022SynchromeshRC} and \cite{Nan2023EnhancingFT,An2023SkillBasedFS} assumes cross-domain setting.
These approaches select few-shot exemplars from the training set based on (a) the semantic similarity with the NL test query \citep{Nan2023EnhancingFT}  or  (b) using Target Similarity Tuning (TST) where the NL queries with similar target programs are selected as exemplars \cite{Poesia2022SynchromeshRC} or (c) selecting similar NL queries as exemplars with their LLM generated `skill' representation, which focuses on program compositions and ignores the surface NL forms \cite{An2023SkillBasedFS}. In addition to TST, Synchromesh \citep{Poesia2022SynchromeshRC} performs constrained semantic decoding (CSD), which reduces the implementation errors in the generated SQLs by ensuring that the generated tokens lead to correct programs following a pre-specified grammar. As constraint decoding is not the focus our approach, we compare our performance using Target Semantic Tuning (TST), without Constraint Semantic Decoding (CSD). All the above similarity based approaches have a reliance on inference-time retrieval of similar few-shot samples from the available data to build a run-time prompt and generate SQL for a test NL query leading to a less efficient solution. As opposed to this, we devise an algorithm to generate a prompt with diverse exemplars generic across test queries in offline fashion, resulting in a more time-efficient solution obviating the need for real time retrieval. Moreover, with further adaptation of this offline prompt with the adapt and decompose techniques, our approach yields better performance than existing similarity based sampling techniques \cite{poesia2022synchromesh,Chang2023HowTP,An2023SkillBasedFS}.

\cite{Nan2023EnhancingFT} defines a `diversity' based sampling method for few-shot exemplar selection, where the samples in the training set are clustered using a combination of continuous representation of the NL queries and discrete representation of SQL counterparts, to pickup the near-centroid samples are few-shots. They showcase that this diversity based sampling method performs better than similarity based sampling, which is further enhanced by similarity-diversity sampling. Our \textit{GP} based sampling technique ensures diversity in the samples, by guaranteeing  complete coverage of SQL operators and maximum converge of database domains.

\subsection{Addressing LLM Memorization Concerns}
\cite{Rajkumar2022EvaluatingTT} have addressed the concerns around possible memorization of existing datasets such as Spider\cite{yu2018spider} by large language models, which are trained on code data. The possibility of memorization arises as the the \textit{Spider Dev} split file (\textit{dev.sql}) resides on Github\footnote{https://github.com/taoyds/spider/tree/master/evaluation\_examples}. However, prompting LLMs with verbatim fragments of this file leads to generations which do not match with the file contents. For example, given a question in the format specified in the file, the table aliasing strategy followed in the generated SQLs does not match with the gold SQLs provided in the file. On the similar lines of \cite{Rajkumar2022EvaluatingTT}, our prompting format of text queries (Explained in Section \ref{sec:DBformat}) is completely different than the format in which NL-SQL pairs are stored in the \textit{Spider Dev} split file. Moreover, to avoid the concerns of memorization, we assess the performance of our pipelines using \textit{Kaggle DBQA} \cite{lee-2021-kaggle-dbqa} dataset, for which the evaluation files are not residing on Github\footnote{https://github.com/chiahsuan156/KaggleDBQA}.  We also showcase with performance improvements with our approach over zero-shot setting on this dataset for distinct LLMs. 

\subsection{Prompts}
In this section we provide the prompts generated by our pipeline including the (i) \textit{GP} which is common across all the \textit{KaggleDBQA} dataset (ii) \textit{DA-GP} for \textit{GeoNeuclear} database. The same prompt template can be used to recreate the prompts for other databases. (iii) \textit{LTMP-GP} which is common across all the \textit{KaggleDBQA} dataset  (iv) \textit{LTMP-DA-GP} for \textit{GeoNeuclear} database. The same prompt template can be used to recreate the prompts for other databases.The \colorbox{yellow}{Yellow} part indicates the few-shot schemas, \colorbox{blue}{Blue} part the test schema and queries and \textcolor{orange}{orange} the domain information.

\onecolumn
\fontsize{10pt}{12pt}\selectfont
\subsubsection{Generic Prompt (GP)} 
Note: PK and FK denote Primarky Key and Foreign Key, respectively, in all the below following schemas. \\
\#\#\# SQLite SQL tables, with their properties: \\
\# \colorbox{yellow}{CREATE TABLE} classroom (building, room\_number, capacity, PK (building, room\_number)) \\
\# \colorbox{yellow}{CREATE TABLE} department (dept\_name, building, budget, PK (dept\_name)) \\
\# \colorbox{yellow}{CREATE TABLE} course (course\_id, title, dept\_name, credits, PK (course\_id), FK (dept\_name) REFERENCES department (dept\_name)) \\
\# \colorbox{yellow}{CREATE TABLE} instructor (ID, name, dept\_name, salary, PK (ID), FK (dept\_name) references department (dept\_name)) \\
\# \colorbox{yellow}{CREATE TABLE} section (course\_id, sec\_id, semester), year, building, room\_number, time\_slot\_id, PK (course\_id, sec\_id, semester, year), FK (course\_id) references course (course\_id), FK (building, room\_number) references classroom (building, room\_number)) \\
\# \colorbox{yellow}{CREATE TABLE} teaches (ID, course\_id, sec\_id, semester, year, PK (ID, course\_id, sec\_id, semester, year), FK (course\_id, sec\_id, semester, year) references section (course\_id, sec\_id, semester, year), FK (ID) references instructor (ID)) \\
\# \colorbox{yellow}{CREATE TABLE} student (ID, name, dept\_name, tot\_cred, PK (ID), FK (dept\_name) references department (dept\_name)) \\
\# \colorbox{yellow}{CREATE TABLE} takes (ID, course\_id, sec\_id, semester, year, grade, PK (ID, course\_id, sec\_id, semester, year), FK (course\_id,sec\_id, semester, year) references section (course\_id, sec\_id, semester, year), FK (ID) references student (ID)) \\
\# \colorbox{yellow}{CREATE TABLE} advisor (s\_ID, i\_ID, PK (s\_ID), FK (i\_ID) references instructor (ID), FK (s\_ID) references student (ID)) \\
\# \colorbox{yellow}{CREATE TABLE} time\_slot (time\_slot\_id, day, start\_hr, start\_min, end\_hr, end\_min, PK (time\_slot\_id, day, start\_hr, start\_min)) \\
\# \colorbox{yellow}{CREATE TABLE} prereq (course\_id, prereq\_id, PK (course\_id, prereq\_id), FK (course\_id) references course (course\_id), FK (prereq\_id) references course (course\_id)) \\
\# \\
\#\#\# Find the buildings which have rooms with capacity more than 50. \\
\textcolor{blue}{SELECT} DISTINCT building \textcolor{blue}{FROM} classroom \textcolor{blue}{WHERE} capacity > 50 \\
\# \\
\#\#\# Find the name and building of the department with the highest budget. \\
\textcolor{blue}{SELECT} dept\_name, building \textcolor{blue}{FROM} department ORDER BY budget DESC LIMIT 1 \\
\# \\
\#\#\# Find the title of courses that have two prerequisites? \\
\textcolor{blue}{SELECT} T1.title \textcolor{blue}{FROM} course AS T1 JOIN prereq AS T2 ON T1.course\_id = T2.course\_id \textcolor{blue}{GROUP BY} T2.course\_id HAVING count(*) = 2 \\
\# \\
\#\#\# How many courses that do not have prerequisite? \\
\textcolor{blue}{SELECT} count(*) \textcolor{blue}{FROM} course \textcolor{blue}{WHERE} course\_id NOT IN (\textcolor{blue}{SELECT} course\_id \textcolor{blue}{FROM} prereq) \\
\# \\
\#\#\# Find the total budgets of the Marketing or Finance department. \\
\textcolor{blue}{SELECT} sum(budget) \textcolor{blue}{FROM} department \textcolor{blue}{WHERE} dept\_name = 'Marketing' OR dept\_name = 'Finance' \\
\# \\
\#\#\# Find the department name of the instructor whose name contains 'Soisalon'. \\
\textcolor{blue}{SELECT} dept\_name \textcolor{blue}{FROM} instructor \textcolor{blue}{WHERE} name LIKE '
\# \\
\#\#\# Find the title of course that is provided by both Statistics and Psychology departments. \\
\textcolor{blue}{SELECT} title \textcolor{blue}{FROM} course \textcolor{blue}{WHERE} dept\_name = 'Statistics' INTERSECT \textcolor{blue}{SELECT} title \textcolor{blue}{FROM} course \textcolor{blue}{WHERE} dept\_name = 'Psychology' \\
\# \\
\#\#\# Find the title of course that is provided by Statistics but not Psychology departments. \\
\textcolor{blue}{SELECT} title \textcolor{blue}{FROM} course \textcolor{blue}{WHERE} dept\_name = 'Statistics' EXCEPT \textcolor{blue}{SELECT} title \textcolor{blue}{FROM} course \textcolor{blue}{WHERE} dept\_name = 'Psychology' \\
\# \\
\#\#\# Find courses that ran in Fall 2009 or in Spring 2010. \\
\textcolor{blue}{SELECT} course\_id \textcolor{blue}{FROM} SECTION \textcolor{blue}{WHERE} semester = 'Fall' AND YEAR = 2009 UNION \textcolor{blue}{SELECT} course\_id \textcolor{blue}{FROM} SECTION \textcolor{blue}{WHERE} semester = 'Spring' AND YEAR = 2010 \\
\# \\
\#\#\# Find the names and average salaries of all departments whose average salary is greater than 42000. \\
\textcolor{blue}{SELECT} dept\_name, AVG(salary) \textcolor{blue}{FROM} instructor \textcolor{blue}{GROUP BY} dept\_name HAVING AVG (salary) > 42000 \\
\# \\
\#\#\# SQLite SQL tables, with their properties: \\
\# \colorbox{yellow}{CREATE TABLE} regions (REGION\_ID, REGION\_NAME, PK (REGION\_ID)) \\
\# \colorbox{yellow}{CREATE TABLE} countries (COUNTRY\_ID, COUNTRY\_NAME, REGION\_ID, PK (COUNTRY\_ID), FK (REGION\_ID) REFERENCES regions (REGION\_ID)) \\
\# \colorbox{yellow}{CREATE TABLE} departments (DEPARTMENT\_ID, DEPARTMENT\_NAME, MANAGER\_ID, LOCATION\_ID, PK (DEPARTMENT\_ID)) \\
\# \colorbox{yellow}{CREATE TABLE} jobs (JOB\_ID, JOB\_TITLE, MIN\_SALARY, MAX\_SALARY, PK (JOB\_ID)) \\
\# \colorbox{yellow}{CREATE TABLE} employees (EMPLOYEE\_ID, FIRST\_NAME, LAST\_NAME, EMAIL, PHONE\_NUMBER, HIRE\_DATE, JOB\_ID, SALARY, COMMISSION\_PCT, MANAGER\_ID, DEPARTMENT\_ID, PK (EMPLOYEE\_ID), FK (DEPARTMENT\_ID) REFERENCES departments(DEPARTMENT\_ID), FK (JOB\_ID) REFERENCES jobs(JOB\_ID)) \\
\# \colorbox{yellow}{CREATE TABLE} job\_history (EMPLOYEE\_ID, START\_DATE, END\_DATE, JOB\_ID, DEPARTMENT\_ID, PK (EMPLOYEE\_ID,START\_DATE), FK (EMPLOYEE\_ID) REFERENCES employees(EMPLOYEE\_ID), FK (DEPARTMENT\_ID) REFERENCES departments(DEPARTMENT\_ID), FK (JOB\_ID) REFERENCES jobs(JOB\_ID)) \\
\# \colorbox{yellow}{CREATE TABLE} locations (LOCATION\_ID, STREET\_ADDRESS, POSTAL\_CODE, CITY, STATE\_PROVINCE, COUNTRY\_ID, PK (LOCATION\_ID), FK (COUNTRY\_ID) REFERENCES countries(COUNTRY\_ID)) \\
\# \\
\#\#\# display job Title, the difference between minimum and maximum salaries for those jobs which max salary within the range 12000 to 18000. \\
\textcolor{blue}{SELECT} job\_title, max\_salary - min\_salary \textcolor{blue}{FROM} jobs \textcolor{blue}{WHERE} max\_salary BETWEEN 12000 AND 18000 \\
\# \\
\#\#\# display the employee ID for each employee and the date on which he ended his previous job. \\
\textcolor{blue}{SELECT} employee\_id, MAX(end\_date) \textcolor{blue}{FROM} job\_history \textcolor{blue}{GROUP BY} employee\_id \\
\# \\
\#\#\# return the smallest salary for every departments. \\
\textcolor{blue}{SELECT} MIN(salary), department\_id \textcolor{blue}{FROM} employees \textcolor{blue}{GROUP BY} department\_id \\
\# \\
\#\#\# display the department id and the total salary for those departments which contains at least two employees. \\
\textcolor{blue}{SELECT} department\_id, SUM(salary) \textcolor{blue}{FROM} employees \textcolor{blue}{GROUP BY} department\_id HAVING count(*) >= 2 \\
\# \\
\#\#\# SQLite SQL tables, with their properties: \\
\# \colorbox{yellow}{CREATE TABLE} Rooms (RoomId PK, roomName, beds, bedType, maxOccupancy, basePrice, decor) \\
\# \colorbox{yellow}{CREATE TABLE} Reservations (Code PK, Room, CheckIn, CheckOut, Rate REAL, LastName, FirstName, Adults, Kids, FK (Room) REFERENCES Rooms(RoomId)) \\
\# \\
\#\#\# List how many times the number of people in the room reached the maximum occupancy of the room. The number of people include adults and kids. \\
\textcolor{blue}{SELECT} count(*) \textcolor{blue}{FROM} Reservations AS T1 JOIN Rooms AS T2 ON T1.Room = T2.RoomId \textcolor{blue}{WHERE} T2.maxOccupancy = T1.Adults + T1.Kids; \\
\# \\
\#\#\# SQLite SQL tables, with their properties: \\
\# \colorbox{yellow}{CREATE TABLE} Attribute\_Definitions (attribute\_id PK, attribute\_name, attribute\_data\_type) \\
\# \colorbox{yellow}{CREATE TABLE} Catalogs (catalog\_id PK, catalog\_name, catalog\_publisher, date\_of\_publication, date\_of\_latest\_revision) \\
\# \colorbox{yellow}{CREATE TABLE} Catalog\_Structure (catalog\_level\_number PK, catalog\_id, catalog\_level\_name, FK (catalog\_id) REFERENCES Catalogs(catalog\_id)) \\
\# \colorbox{yellow}{CREATE TABLE} Catalog\_Contents (catalog\_entry\_id PK, catalog\_level\_number, parent\_entry\_id, previous\_entry\_id, next\_entry\_id, catalog\_entry\_name, product\_stock\_number, price\_in\_dollars, price\_in\_euros, price\_in\_pounds, capacity, length, height, width, FK (catalog\_level\_number) REFERENCES Catalog\_Structure(catalog\_level\_number)) \\
\# \colorbox{yellow}{CREATE TABLE} Catalog\_Contents\_Additional\_Attributes (catalog\_entry\_id, catalog\_level\_number, attribute\_id, attribute\_value, FK (catalog\_entry\_id) REFERENCES Catalog\_Contents(catalog\_entry\_id), FK (catalog\_level\_number) REFERENCES Catalog\_Structure(catalog\_level\_number)) \\
\# \\
\#\#\# Find the names of the products with length smaller than 3 or height greater than 5. \\
\textcolor{blue}{SELECT} catalog\_entry\_name \textcolor{blue}{FROM} catalog\_contents \textcolor{blue}{WHERE} LENGTH < 3 OR width > 5 \\
\colorbox{cyan}{\#} \\
\colorbox{cyan}{\#\#\# SQLite SQL tables, with their properties:} \\
\colorbox{cyan}{\# CREATE TABLE nuclear\_power\_plants (Id, Name, Latitude, Longitude, Country, Status, ReactorType,} \\
\colorbox{cyan}{ReactorModel, ConstructionStartAt, OperationalFrom, OperationalTo, Capacity, LastUpdatedAt, Source)} \\
\colorbox{cyan}{\#} \\
\colorbox{cyan}{\#\#\# Which country has the most capacities of nuclear power plants?} \\
\colorbox{cyan}{\textcolor{blue}{SELECT}}
\\

\subsubsection{Domain Adapted - Generic Prompt (DA-GP) (Stage-3)} 

\#\#\# SQLite SQL tables, with their properties: \\
\# \\
\# \colorbox{yellow}{CREATE TABLE nuclear\_power\_plants} (Id, Name, Latitude, Longitude, Country, Status, ReactorType, ReactorModel, ConstructionStartAt, OperationalFrom, OperationalTo, Capacity, LastUpdatedAt, Source) \\
Columns in nuclear\_power\_plants with examples in each column and descriptions wherever required: \\
\textcolor{orange}{Id}: 572, 560, 258, 433. \\
\textcolor{orange}{Name}: Ågesta, Turkey Point\-4, Oskarshamn\-2, Ningde\-4. \\
\textcolor{orange}{Latitude}: 55.084000, 55.604000, 41.188000, 45.800000. Description: latitude in decimal format \\
\textcolor{orange}{Longitude}: \-77.311000, 66.790000, 9.393000, 0.845000. Description: longitude in decimal format \\
\textcolor{orange}{Country}: Canada, Germany, Taiwan, Province of China, Italy. \\
\textcolor{orange}{Status}: Planned, Cancelled Construction, Under Construction, Suspended Construction. \\
\textcolor{orange}{ReactorType}: HWGCR, GCR, LWGR, HTGR. \\
\textcolor{orange}{ReactorModel}: Konvoi, VVER V-320, WH 2LP (DRYAMB), PHWR KWU. \\
\textcolor{orange}{ConstructionStartAt}: 1977-02-01, 1968-05-18, 1965-04-12, 1972-11-01. Description: date when nuclear power plant construction was started \\
\textcolor{orange}{OperationalFrom}: 2015-06-05, 1977-03-13, 1986-04-10, 1989-09-30. Description: date when nuclear power plant became operational (also known as commercial operation date) \\
\textcolor{orange}{OperationalTo}: 2011-05-19, 2004-06-29, 1992-05-27, 2015-04-30. Description: date when nuclear power plant was shutdown (also known as permanent shutdown date) \\
\textcolor{orange}{Capacity}: 1092, 125, 535, 1307. Description: nuclear power plant capacity (design net capacity in MWe) \\
\textcolor{orange}{LastUpdatedAt}: 2015-05-24T04:50:59+03:00, 2015-05-24T04:51:11+03:00, 2017-02-10T23:58:48+02:00, 2018-03-10T13:41:49+02:00. Description: date and time when information was last updated \\
\textcolor{orange}{Source}: WNA/wikipedia/IAEA, wikipedia, WNA, WNA/IAEA/GEO. Description: source of the information \\
\# \\
\#\#\# Find the latitudes of nuclear power plants with capacity more than 50. \\
\textcolor{blue}{SELECT} \textcolor{blue}{DISTINCT} Latitude \textcolor{blue}{FROM} nuclear\_power\_plants \textcolor{blue}{WHERE} Capacity > 50; \\
\# \\
\#\#\# Find the country and status of the nuclear power plant with the highest capacity. \\
\textcolor{blue}{SELECT} Country, Status \textcolor{blue}{FROM} nuclear\_power\_plants \textcolor{blue}{ORDER} BY Capacity DESC LIMIT 1; \\
\# \\
\#\#\# Find the name of nuclear power plants that have two entries in the database? \\
\textcolor{blue}{SELECT} T1.Name \textcolor{blue}{FROM} nuclear\_power\_plants AS T1 JOIN nuclear\_power\_plants AS T2 ON T1.Id = T2.Id GROUP BY T2.Id HAVING count(*) = 2; \\
\# \\
\#\#\# How many nuclear power plants do not have a prerequisite? \\
\textcolor{blue}{SELECT} COUNT(*) \textcolor{blue}{FROM} nuclear\_power\_plants \textcolor{blue}{WHERE} Id NOT IN (\textcolor{blue}{SELECT} Id \textcolor{blue}{FROM} prereq) \\
\# \\
\#\#\# Find the total capacity of the nuclear power plants named Marketing or Finance. \\
\textcolor{blue}{SELECT} sum(Capacity) \textcolor{blue}{FROM} nuclear\_power\_plants \textcolor{blue}{WHERE} Name = 'Marketing' OR Name = 'Finance' \\
\# \\
\#\#\# Find the country associated with the nuclear power plant whose name contains 'Soisalon'. \\
\textcolor{blue}{SELECT} Country \textcolor{blue}{FROM} nuclear\_power\_plants \textcolor{blue}{WHERE} Name LIKE '\%Soisalon\%' \\
\# \\
\#\#\# Find the name of nuclear power plants that are located in both Statistics and Psychology countries. \\
\textcolor{blue}{SELECT} Name \textcolor{blue}{FROM} nuclear\_power\_plants \textcolor{blue}{WHERE} Country = 'Statistics' INTERSECT \textcolor{blue}{SELECT} Name \textcolor{blue}{FROM} nuclear\_power\_plants \textcolor{blue}{WHERE} Country = 'Psychology' \\
\# \\
\#\#\# Find the name of nuclear power plants that are located in Statistics but not Psychology countries. \\
\textcolor{blue}{SELECT} Name \textcolor{blue}{FROM} nuclear\_power\_plants \textcolor{blue}{WHERE} Country = 'Statistics' EXCEPT \textcolor{blue}{SELECT} Name \textcolor{blue}{FROM} nuclear\_power\_plants \textcolor{blue}{WHERE} Country = 'Psychology' \\
\# \\
\#\#\# Find the Ids of nuclear power plants that were constructed in Fall 2009 or in Spring 2010. \\
\textcolor{blue}{SELECT} Id \textcolor{blue}{FROM} nuclear\_power\_plants \textcolor{blue}{WHERE} ConstructionStartAt = 'Fall' AND LastUpdatedAt = 2009 UNION \textcolor{blue}{SELECT} Id \textcolor{blue}{FROM} nuclear\_power\_plants \textcolor{blue}{WHERE} ConstructionStartAt = 'Spring' AND LastUpdatedAt = 2010 \\
\# \\
\#\#\# Find the countries and average capacities of all nuclear power plants whose average capacity is greater than 42000. \\
\textcolor{blue}{SELECT} Country, AVG (Capacity) \textcolor{blue}{FROM} nuclear\_power\_plants GROUP BY Country HAVING AVG (Capacity) > 42000 \\
\# \\
\#\#\# display the Name of the nuclear power plant and the difference between its capacity and the year it was constructed for those plants which capacity is within the range 12000 to 18000. \\
\textcolor{blue}{SELECT} Name, Capacity - ConstructionStartAt \textcolor{blue}{FROM} nuclear\_power\_plants \textcolor{blue}{WHERE} Capacity BETWEEN 12000 AND 18000 \\
\# \\
\#\#\# display the ID for each nuclear power plant and the date on which it stopped operating. \\
\textcolor{blue}{SELECT} Id, MAX(OperationalTo) \textcolor{blue}{FROM} nuclear\_power\_plants GROUP BY Id \\
\# \\
\#\#\# return the smallest capacity for each nuclear power plant. \\
\textcolor{blue}{SELECT} MIN(Capacity), Id \textcolor{blue}{FROM} nuclear\_power\_plants GROUP BY Id \\
\# \\
\#\#\# display the id and the total capacity for those nuclear power plants which have at least two reactors. \\
\textcolor{blue}{SELECT} Id, SUM(Capacity) \textcolor{blue}{FROM} nuclear\_power\_plants GROUP BY Id HAVING count(*) >= 2 \\
\# \\
\#\#\# Count how many times the capacity of a nuclear power plant is equal to the sum of its status and reactor type. \\
\textcolor{blue}{SELECT} COUNT(*) \textcolor{blue}{FROM} nuclear\_power\_plants AS T1 JOIN nuclear\_power\_plants AS T2 ON T1.Id = T2.Id \textcolor{blue}{WHERE} T2.Capacity = T1.Status + T1.ReactorType; \\
\# \\
\#\#\# Find the names of the nuclear power plants with capacity smaller than 3 or capacity greater than 5. \\
\textcolor{blue}{SELECT} Name \textcolor{blue}{FROM} nuclear\_power\_plants \textcolor{blue}{WHERE} Capacity < 3 OR Capacity > 5 \\
\colorbox{cyan}{\#} \\
\colorbox{cyan}{\#\#\# Which country has the most capacities of nuclear power plants?} \\
\colorbox{cyan}{\textcolor{blue}{SELECT} }\\
\\
\subsubsection{Least-to-Most Prompting - Generic Prompt (LTMP-GP)}
\textcolor{red}{Note: PK and FK denote Primary Key and Foreign Key, respectively, in all the below following schemas.} \\
\#\#\# SQLite SQL tables, with their properties: \\
\colorbox{yellow}{\#CREATE TABLE} classroom (building, room\_number, capacity, PK (building, room\_number)) \\
\colorbox{yellow}{\#CREATE TABLE} department (dept\_name, building, budget, PK (dept\_name)) \\
\colorbox{yellow}{\#CREATE TABLE} course (course\_id, title, dept\_name, credits, PK (course\_id), FK (dept\_name) REFERENCES department (dept\_name)) \\
\colorbox{yellow}{\#CREATE TABLE} instructor (ID, name, dept\_name, salary, PK (ID), FK (dept\_name) references department (dept\_name)) \\
\colorbox{yellow}{\#CREATE TABLE} section (course\_id, sec\_id, semester), year, building, room\_number, time\_slot\_id, PK (course\_id, sec\_id, semester, year), FK (course\_id) references course (course\_id), FK (building, room\_number) references classroom (building, room\_number)) \\
\colorbox{yellow}{\#CREATE TABLE} teaches (ID, course\_id, sec\_id, semester, year, PK (ID, course\_id, sec\_id, semester, year), FK (course\_id, sec\_id, semester, year) references section (course\_id, sec\_id, semester, year), FK (ID) references instructor (ID)) \\
\colorbox{yellow}{\#CREATE TABLE} student (ID, name, dept\_name, tot\_cred, PK (ID), FK (dept\_name) references department (dept\_name)) \\
\colorbox{yellow}{\#CREATE TABLE} takes (ID, course\_id, sec\_id, semester, year, grade, PK (ID, course\_id, sec\_id, semester, year), FK (course\_id,sec\_id, semester, year) references section (course\_id, sec\_id, semester, year), FK (ID) references student (ID)) \\
\colorbox{yellow}{\#CREATE TABLE} advisor (s\_ID, i\_ID, PK (s\_ID), FK (i\_ID) references instructor (ID), FK (s\_ID) references student (ID)) \\
\colorbox{yellow}{\#CREATE TABLE} time\_slot (time\_slot\_id, day, start\_hr, start\_min, end\_hr, end\_min, PK (time\_slot\_id, day, start\_hr, start\_min)) \\
\colorbox{yellow}{\#CREATE TABLE} prereq (course\_id, prereq\_id, PK (course\_id, prereq\_id), FK (course\_id) references course (course\_id), FK (prereq\_id) references course (course\_id)) \\
\# \\
Q: Find the buildings which have rooms with capacity more than 50. \\
sub-questions:[Find the buildings, which have rooms with capacity more than 50.] \\
Intermediate representation: ['select distinct classroom.building', 'select where classroom.capacity > 50'] \\
\colorbox{yellow}{A:} Lets think step by step. To get the SQL using the intermediate representations, we combine them to form: \\
\colorbox{yellow}{SQL:} \textcolor{blue}{SELECT} DISTINCT building \textcolor{blue}{FROM} classroom \textcolor{blue}{WHERE} capacity > 50 \\
\# \\
Q: Find the name and building of the department with the highest budget. \\
sub-questions:[Find the name and building of the department, with the highest budget.] \\
Intermediate representation: ['select department.dept\_name, department.building', 'select order by department.budge desc limit 1'] \\
\colorbox{yellow}{A:} Lets think step by step. To get the SQL using the intermediate representations, we combine them to form: \\
\colorbox{yellow}{SQL:} \textcolor{blue}{SELECT} dept\_name, building \textcolor{blue}{FROM} department \textcolor{blue}{ORDER BY} budget DESC LIMIT 1 \\
\# \\
Q: Find the title of courses that have two prerequisites? \\
sub-questions:[Find the title of courses, that have two prerequisites?] \\
Intermediate representation: ['select course.title', 'select where count(prereq.*)=2 group by prereq.course\_id'] \\
\colorbox{yellow}{A:} Lets think step by step. To get the SQL using the intermediate representations, we combine them to form: \\
\colorbox{yellow}{SQL:} \textcolor{blue}{SELECT} T1.title \textcolor{blue}{FROM} course AS T1 JOIN prereq AS T2 ON T1.course\_id = T2.course\_id \textcolor{blue}{GROUP BY} T2.course\_id HAVING count(*) = 2 \\
\# \\
Q: How many courses that do not have prerequisite? \\
sub-questions:[How many courses, that do not have prerequisite?] \\
Intermediate representation: ['select count(Courses.*)', 'select where @.@ not in prereq.course\_id'] \\
\colorbox{yellow}{A:} Lets think step by step. To get the SQL using the intermediate representations, we combine them to form: \\
\colorbox{yellow}{SQL:} \textcolor{blue}{SELECT} count(*) \textcolor{blue}{FROM} course \textcolor{blue}{WHERE} course\_id NOT IN (\textcolor{blue}{SELECT} course\_id \textcolor{blue}{FROM} prereq) \\
\# \\
Q: Find the total budgets of the Marketing or Finance Department. \\
sub-questions:[Find the total budgets of the Marketing or Finance Department.] \\
Intermediate representation: ['select sum(department.budget) where department.dept\_name = "Marketing" or department.dept\_name="Finance"'] \\
\colorbox{yellow}{A:} Lets think step by step. To get the SQL using the intermediate representations, we combine them to form: \\
\colorbox{yellow}{SQL:} \textcolor{blue}{SELECT} sum(budget) \textcolor{blue}{FROM} department \textcolor{blue}{WHERE} dept\_name = 'Marketing' OR dept\_name = 'Finance' \\
\# \\
Q: Find the department name of the instructor whose name contains 'Soisalon'. \\
sub-questions:[Find the department name of the instructor, whose name contains 'Soisalon'.] \\
Intermediate representation: ['select instructor.dept\_name', 'select where instructor.name like "\%Soisalon\%"'] \\
\colorbox{yellow}{A:} Lets think step by step. To get the SQL using the intermediate representations, we combine them to form: \\
\colorbox{yellow}{SQL:} \textcolor{blue}{SELECT} dept\_name \textcolor{blue}{FROM} instructor \textcolor{blue}{WHERE} name LIKE '\%Soisalon\%' \\
\# \\
Q: Find the title of course that is provided by both Statistics and Psychology departments. \\
sub-questions:[Find the title of course, that is provided by both Statistics and Psychology departments.] \\
Intermediate representation: ['select course.title', 'select where course.dept\_name="Statistics" and course.dept\_name="Psychology"'] \\
\colorbox{yellow}{A:} Lets think step by step. To get the SQL using the intermediate representations, we combine them to form: \\
\colorbox{yellow}{SQL:} \textcolor{blue}{SELECT} title \textcolor{blue}{FROM} course \textcolor{blue}{WHERE} dept\_name = 'Statistics' INTERSECT \textcolor{blue}{SELECT} title \textcolor{blue}{FROM} course \textcolor{blue}{WHERE} dept\_name = 'Psychology' \\
\# \\
Q: Find the title of course that is provided by Statistics but not Psychology departments. \\
sub-questions:[Find the title of course, that is provided by Statistics, but not Psychology departments.] \\
Intermediate representation: ['select course.title', 'select where course.dept\_name="Statistics"', 'select where course.dept\_name!="Psychology"'] \\
\colorbox{yellow}{A:} Lets think step by step. To get the SQL using the intermediate representations, we combine them to form: \\
\colorbox{yellow}{SQL:} \textcolor{blue}{SELECT} title \textcolor{blue}{FROM} course \textcolor{blue}{WHERE} dept\_name = 'Statistics' EXCEPT \textcolor{blue}{SELECT} title \textcolor{blue}{FROM} course \textcolor{blue}{WHERE} dept\_name = 'Psychology' \\
\# \\
Q: Find courses that ran in Fall 2009 or in Spring 2010. \\
sub-questions:[Find courses, that ran in Fall 2009 or, in Spring 2010.] \\
Intermediate representation: ['select section.course\_id', 'select where section.semester="Fall" and section.year=2009', 'select where section.semester="Spring" and section.year=2010'] \\
\colorbox{yellow}{A:} Lets think step by step. To get the SQL using the intermediate representations, we combine them to form: \\
\colorbox{yellow}{SQL:} \textcolor{blue}{SELECT} course\_id \textcolor{blue}{FROM} SECTION \textcolor{blue}{WHERE} semester = 'Fall' AND YEAR = 2009 UNION \textcolor{blue}{SELECT} course\_id \textcolor{blue}{FROM} SECTION \textcolor{blue}{WHERE} semester = 'Spring' AND YEAR = 2010 \\
\# \\
Q: Find the names and average salaries of all departments whose average salary is greater than 42000. \\
sub-questions:[Find the names and average salaries of all departments, whose average salary is greater than 42000.] \\
Intermediate representation: ['select instructor.dept\_name, avg(isntructor.slary) group by instructor.dept\_name', 'select where avg(instructor.salary) > 42000'] \\
\colorbox{yellow}{A:} Lets think step by step. To get the SQL using the intermediate representations, we combine them to form: \\
\colorbox{yellow}{SQL:} \textcolor{blue}{SELECT} dept\_name, AVG(salary) \textcolor{blue}{FROM} instructor \textcolor{blue}{GROUP BY} dept\_name HAVING AVG (salary) > 42000 \\
\# \\
\#\#\# SQLite SQL tables, with their properties: \\
\colorbox{yellow}{\#CREATE TABLE} regions (REGION\_ID, REGION\_NAME, PK (REGION\_ID)) \\
\colorbox{yellow}{\#CREATE TABLE} countries (COUNTRY\_ID, COUNTRY\_NAME, REGION\_ID, PK (COUNTRY\_ID), FK (REGION\_ID) REFERENCES regions (REGION\_ID)) \\
\colorbox{yellow}{\#CREATE TABLE} departments (DEPARTMENT\_ID, DEPARTMENT\_NAME, MANAGER\_ID, LOCATION\_ID, PK (DEPARTMENT\_ID)) \\
\colorbox{yellow}{\#CREATE TABLE} jobs (JOB\_ID, JOB\_TITLE, MIN\_SALARY, MAX\_SALARY, PK (JOB\_ID)) \\
\colorbox{yellow}{\#CREATE TABLE} employees (EMPLOYEE\_ID, FIRST\_NAME, LAST\_NAME, EMAIL, PHONE\_NUMBER, HIRE\_DATE, JOB\_ID, SALARY, COMMISSION\_PCT, MANAGER\_ID, DEPARTMENT\_ID, PK (EMPLOYEE\_ID), FK (DEPARTMENT\_ID) REFERENCES departments(DEPARTMENT\_ID), FK (JOB\_ID) REFERENCES jobs(JOB\_ID)) \\
\colorbox{yellow}{\#CREATE TABLE} job\_history (EMPLOYEE\_ID, START\_DATE, END\_DATE, JOB\_ID, DEPARTMENT\_ID, PK (EMPLOYEE\_ID,START\_DATE), FK (EMPLOYEE\_ID) REFERENCES employees(EMPLOYEE\_ID), FK (DEPARTMENT\_ID) REFERENCES departments(DEPARTMENT\_ID), FK (JOB\_ID) REFERENCES jobs(JOB\_ID)) \\
\colorbox{yellow}{\#CREATE TABLE} locations (LOCATION\_ID, STREET\_ADDRESS, POSTAL\_CODE, CITY, STATE\_PROVINCE, COUNTRY\_ID, PK (LOCATION\_ID), FK (COUNTRY\_ID) REFERENCES countries(COUNTRY\_ID)) \\
\# \\
Q: display job Title, the diffrence between minimum and maximum salaries for those jobs which max salary within the range 12000 to 18000. \\
sub-questions:[display job Title, the diffrence between minimum and maximum salaries for those jobs, which max salary within the range 12000 to 18000.] \\
Intermediate representation: ['select jobs.JOB\_TITLE, jobs.MAX\_SALARY - jobs.MIN\_SALARY', 'select where jobs.MAX\_SALARY between 12000 and 18000'] \\
\colorbox{yellow}{A:} Lets think step by step. To get the SQL using the intermediate representations, we combine them to form: \\
\colorbox{yellow}{SQL:} \textcolor{blue}{SELECT} job\_title, max\_salary - min\_salary \textcolor{blue}{FROM} jobs \textcolor{blue}{WHERE} max\_salary BETWEEN 12000 AND 18000 \\
\# \\
Q: display the employee ID for each employee and the date on which he ended his previous job. \\
sub-questions:[display the employee ID for each employee and the date, on which he ended his previous job.] \\
Intermediate representation: ['select job\_history.EMPLOYEE\_ID, max(job\_history.END\_DATE) group by job\_history.EMPLOYEE\_ID', 'select extra max (job\_history.END\_DATE)'] \\
\colorbox{yellow}{A:} Lets think step by step. To get the SQL using the intermediate representations, we combine them to form: \\
\colorbox{yellow}{SQL:} \textcolor{blue}{SELECT} employee\_id, MAX(end\_date) \textcolor{blue}{FROM} job\_history \textcolor{blue}{GROUP BY} employee\_id \\
\# \\
Q: return the smallest salary for every departments. \\
sub-questions:[return the smallest salary for every departments.] \\
Intermediate representation: ['select min(employees.SALARY), employees.DEPARTMENT\_ID group by employees.DEPARTMENT\_ID'] \\
\colorbox{yellow}{A:} Lets think step by step. To get the SQL using the intermediate representations, we combine them to form: \\
\colorbox{yellow}{SQL:} \textcolor{blue}{SELECT} MIN(salary), department\_id \textcolor{blue}{FROM} employees \textcolor{blue}{GROUP BY} department\_id \\
\# \\
Q: display the department id and the total salary for those department which contains at least two employees. \\
sub-questions:[display the department id and the total salary for those department, which contains at least two employees.] \\
Intermediate representation: ['select employees.DEPARTEMENT\_ID, sum(employees.SALARY)', 'select where count (employees.*)>=2 group by employees.DEPARTMENT\_ID'] \\
\colorbox{yellow}{A:} Lets think step by step. To get the SQL using the intermediate representations, we combine them to form: \\
\colorbox{yellow}{SQL:} \textcolor{blue}{SELECT} department\_id, SUM(salary) \textcolor{blue}{FROM} employees \textcolor{blue}{GROUP BY} department\_id HAVING count(*) >= 2 \\
\# \\
\#\#\# SQLite SQL tables, with their properties: \\
\colorbox{yellow}{\#CREATE TABLE} Rooms (RoomId PK, roomName, beds, bedType, maxOccupancy, basePrice, decor) \\
\colorbox{yellow}{\#CREATE TABLE} Reservations (Code PK, Room, CheckIn, CheckOut, Rate REAL, LastName, FirstName, Adults, Kids, FK (Room) REFERENCES Rooms(RoomId)) \\
\# \\
Q: List how many times the number of people in the room reached the maximum occupancy of the room. The number of people include adults and kids. \\
sub-questions:[List how many times the number of people in the room, the maximum occupancy of the room., The number of people include adults and kids.] \\
Intermediate representation: ['select count(Resrevations.*)', 'select where Rooms.maxOccupancy=Reservations.Adults','select extra Reservations.Adults'] \\
\colorbox{yellow}{A:} Lets think step by step. To get the SQL using the intermediate representations, we combine them to form: \\
\colorbox{yellow}{SQL:} \textcolor{blue}{SELECT} count(*) \textcolor{blue}{FROM} Reservations AS T1 JOIN Rooms AS T2 ON T1.Room = T2.RoomId \textcolor{blue}{WHERE} T2.maxOccupancy = T1.Adults + T1.Kids; \\
\# \\
\#\#\# SQLite SQL tables, with their properties: \\
\colorbox{yellow}{\#CREATE TABLE} Attribute\_Definitions (attribute\_id PK, attribute\_name, attribute\_data\_type) \\
\colorbox{yellow}{\#CREATE TABLE} Catalogs (catalog\_id PK, catalog\_name, catalog\_publisher, date\_of\_publication, date\_of\_latest\_revision) \\
\colorbox{yellow}{\#CREATE TABLE} Catalog\_Structure (catalog\_level\_number PK, catalog\_id, catalog\_level\_name, FK (catalog\_id) REFERENCES Catalogs(catalog\_id)) \\
\colorbox{yellow}{\#CREATE TABLE} Catalog\_Contents (catalog\_entry\_id PK, catalog\_level\_number, parent\_entry\_id, previous\_entry\_id, next\_entry\_id, catalog\_entry\_name, product\_stock\_number, price\_in\_dollars, price\_in\_euros, price\_in\_pounds, capacity, length, height, width, FK (catalog\_level\_number) REFERENCES Catalog\_Structure(catalog\_level\_number)) \\
\colorbox{yellow}{\#CREATE TABLE} Catalog\_Contents\_Additional\_Attributes (catalog\_entry\_id, catalog\_level\_number, attribute\_id, attribute\_value, FK (catalog\_entry\_id) REFERENCES Catalog\_Contents(catalog\_entry\_id), FK (catalog\_level\_number) REFERENCES Catalog\_Structure(catalog\_level\_number)) \\
\# \\
Q: Find the names of the products with length smaller than 3 or height greater than 5. \\
sub-questions:[Find the names of the products, with length smaller than 3 or, height greater than 5.] \\
Intermediate representation: ['select Catalog\_Contents.catalog\_entry\_name, 'select where Catalog\_Contents.length < 3','select where Catalog\_Contents.width > 5'] \\
\colorbox{yellow}{A:} Lets think step by step. To get the SQL using the intermediate representations, we combine them to form: \\
\colorbox{yellow}{SQL:} \textcolor{blue}{SELECT} catalog\_entry\_name \textcolor{blue}{FROM} catalog\_contents \textcolor{blue}{WHERE} LENGTH < 3 OR width > 5 \\
\colorbox{cyan}{\#} \\
\colorbox{cyan}{\#\#\# SQLite SQL tables, with their properties:} \\
\colorbox{cyan}{\# CREATE TABLE nuclear\_power\_plants (Id, Name, Latitude, Longitude, Country, Status, ReactorType,} \\
\colorbox{cyan}{ReactorModel, ConstructionStartAt, OperationalFrom, OperationalTo, Capacity, LastUpdatedAt, Source)} \\
\colorbox{cyan}{\#} \\
\colorbox{cyan}{Q: Which country has the most capacities of nuclear power plants?} \\
\colorbox{cyan}{A:} \\
\\
\subsection{Least-to-Most Prompting - Domain Adapted - Generic Prompt (LTMP-DA-GP)}
\#\#\# SQLite SQL tables, with their properties: \\
\# \\
\colorbox{yellow}{\# CREATE TABLE nuclear\_power\_plants} (Id, Name, Latitude, Longitude, Country, Status, ReactorType, ReactorModel, ConstructionStartAt, OperationalFrom, OperationalTo, Capacity, LastUpdatedAt, Source) \\
Columns in nuclear\_power\_plants with examples in each column and descriptions wherever required: \\
\textcolor{orange}{Id}: 572, 560, 258, 433. \\
\textcolor{orange}{Name}: Ågesta, Turkey Point-4, Oskarshamn-2, Ningde-4. \\
\textcolor{orange}{Latitude}: 55.084000, 55.604000, 41.188000, 45.800000. Description: latitude in decimal format \\
\textcolor{orange}{Longitude}: -77.311000, 66.790000, 9.393000, 0.845000. Description: longitude in decimal format \\
\textcolor{orange}{Country}: Canada, Germany, Taiwan, Province of China, Italy. \\
\textcolor{orange}{Status}: Planned, Cancelled Construction, Under Construction, Suspended Construction. \\
\textcolor{orange}{ReactorType}: HWGCR, GCR, LWGR, HTGR. \\
\textcolor{orange}{ReactorModel}: Konvoi, VVER V-320, WH 2LP (DRYAMB), PHWR KWU. \\
\textcolor{orange}{ConstructionStartAt}: 1977-02-01, 1968-05-18, 1965-04-12, 1972-11-01. Description: date when nuclear power plant construction was started \\
\textcolor{orange}{OperationalFrom}: 2015-06-05, 1977-03-13, 1986-04-10, 1989-09-30. Description: date when nuclear power plant became operational (also known as commercial operation date) \\
\textcolor{orange}{OperationalTo}: 2011-05-19, 2004-06-29, 1992-05-27, 2015-04-30. Description: date when nuclear power plant was shutdown (also known as permanent shutdown date) \\
\textcolor{orange}{Capacity}: 1092, 125, 535, 1307. Description: nuclear power plant capacity (design net capacity in MWe) \\
\textcolor{orange}{LastUpdatedAt}: 2015-05-24T04:50:59+03:00, 2015-05-24T04:51:11+03:00, 2017-02-10T23:58:48+02:00, 2018-03-10T13:41:49+02:00. Description: date and time when information was last updated \\
\textcolor{orange}{Source}: WNA/wikipedia/IAEA, wikipedia, WNA, WNA/IAEA/GEO. Description: source of the information \\
\# \\
Q: Find the latitudes of nuclear power plants with capacity more than 50. \\
sub-questions:['Find the latitudes of nuclear power plants', 'with capacity more than 50.'] \\
Intermediate representation: ['select distinct nuclear\_power\_plants.Latitude', 'select where nuclear\_power\_plants.Capacity > 50'] \\
\colorbox{yellow}{A:} Lets think step by step. To get the SQL using the intermediate representations, we combine them to form: \\
\colorbox{yellow}{SQL:} [ \textcolor{blue}{SELECT} DISTINCT Latitude \textcolor{blue}{FROM} nuclear\_power\_plants \textcolor{blue}{WHERE} Capacity > 50 ] \\
\# \\
Q: Find the country and status of the nuclear power plant with the highest capacity. \\
sub-questions:['Find the country and status of the nuclear power plant', 'with the highest capacity.'] \\
Intermediate representation: ['select nuclear\_power\_plants.Country, nuclear\_power\_plants.Status', 'select order by nuclear\_power\_plants.Capacity desc limit 1'] \\
\colorbox{yellow}{A:} Lets think step by step. To get the SQL using the intermediate representations, we combine them to form: \\
\colorbox{yellow}{SQL:} [ \textcolor{blue}{SELECT} Country, Status \textcolor{blue}{FROM} nuclear\_power\_plants \textcolor{blue}{ORDER BY} Capacity DESC LIMIT 1 ] \\
\# \\
Q: Find the name of nuclear power plants that have two entries in the database? \\
sub-questions:['Find the name of nuclear power plants', 'that have two entries in the database?'] \\
Intermediate representation: ['select nuclear\_power\_plants.Name', 'select where count(nuclear\_power\_plants.*)=2 group by nuclear\_power\_plants.Id'] \\
\colorbox{yellow}{A:} Lets think step by step. To get the SQL using the intermediate representations, we combine them to form: \\
\colorbox{yellow}{SQL:} [ \textcolor{blue}{SELECT} T1.Name \textcolor{blue}{FROM} nuclear\_power\_plants AS T1 JOIN nuclear\_power\_plants AS T2 ON T1.Id = T2.Id \textcolor{blue}{GROUP BY} T2.Id HAVING count(*) = 2 ] \\
\# \\
Q: How many nuclear power plants do not have a prerequisite? \\
sub-questions:['How many nuclear power plants', 'do not have a prerequisite?'] \\
Intermediate representation: ['select count(nuclear\_power\_plants.*)', 'select where @.@ not in prereq.Id'] \\
\colorbox{yellow}{A:} Lets think step by step. To get the SQL using the intermediate representations, we combine them to form: \\
\colorbox{yellow}{SQL:} [ \textcolor{blue}{SELECT} COUNT(*) \textcolor{blue}{FROM} nuclear\_power\_plants \textcolor{blue}{WHERE} Id NOT IN (\textcolor{blue}{SELECT} Id \textcolor{blue}{FROM} prereq) ] \\
\# \\
Q: Find the total capacity of the nuclear power plants named Marketing or Finance. \\
sub-questions:['Find the total capacity of the nuclear power plants named Marketing or Finance.'] \\
Intermediate representation: ['select sum(nuclear\_power\_plants.Capacity) where nuclear\_power\_plants.Name = "Marketing" or nuclear\_power\_plants.Name="Finance"'] \\
\colorbox{yellow}{A:} Lets think step by step. To get the SQL using the intermediate representations, we combine them to form: \\
\colorbox{yellow}{SQL:} [ \textcolor{blue}{SELECT} sum(Capacity) \textcolor{blue}{FROM} nuclear\_power\_plants \textcolor{blue}{WHERE} Name = 'Marketing' OR Name = 'Finance' ] \\
\# \\
Q: Find the country associated with the nuclear power plant whose name contains 'Soisalon'. \\
sub-questions:['Find the country associated with the nuclear power plant', 'whose name contains "Soisalon"'.] \\
Intermediate representation: ['select nuclear\_power\_plants.Country', 'select where nuclear\_power\_plants.name like "\%Soisalon\%"'] \\
\colorbox{yellow}{A:} Lets think step by step. To get the SQL using the intermediate representations, we combine them to form: \\
\colorbox{yellow}{SQL:} [ \textcolor{blue}{SELECT} Country \textcolor{blue}{FROM} nuclear\_power\_plants \textcolor{blue}{WHERE} Name LIKE '\%Soisalon\%' ] \\
\# \\
Q: Find the name of nuclear power plants that are located in both Statistics and Psychology countries. \\
sub-questions:['Find the name of nuclear power plants', 'that are located in both Statistics and Psychology countries.'] \\
Intermediate representation: ['select nuclear\_power\_plants.Name', 'select where nuclear\_power\_plants.Country="Statistics" and nuclear\_power\_plants.Country="Psychology"'] \\
\colorbox{yellow}{A:} Lets think step by step. To get the SQL using the intermediate representations, we combine them to form: \\
\colorbox{yellow}{SQL:} [ \textcolor{blue}{SELECT} Name \textcolor{blue}{FROM} nuclear\_power\_plants \textcolor{blue}{WHERE} Country = 'Statistics' INTERSECT \textcolor{blue}{SELECT} Name \textcolor{blue}{FROM} nuclear\_power\_plants \textcolor{blue}{WHERE} Country = 'Psychology' ] \\
\# \\
Q: Find the name of nuclear power plants that are located in Statistics but not Psychology countries. \\
sub-questions:['Find the name of nuclear power plants', 'that are located in Statistics but not Psychology countries.'] \\
Intermediate representation: ['select nuclear\_power\_plants.Name', 'select where nuclear\_power\_plants.Country="Statistics"', 'select where nuclear\_power\_plants.Country!="Psychology"'] \\
\colorbox{yellow}{A:} Lets think step by step. To get the SQL using the intermediate representations, we combine them to form: \\
\colorbox{yellow}{SQL:} [ \textcolor{blue}{SELECT} Name \textcolor{blue}{FROM} nuclear\_power\_plants \textcolor{blue}{WHERE} Country = 'Statistics' EXCEPT \textcolor{blue}{SELECT} Name \textcolor{blue}{FROM} nuclear\_power\_plants \textcolor{blue}{WHERE} Country = 'Psychology' ] \\
\# \\
Q: Find the Ids of nuclear power plants that were constructed in Fall 2009 or in Spring 2010. \\
sub-questions:['Find the Ids of nuclear power plants', 'that were constructed in Fall 2009 or, in Spring 2010.'] \\
Intermediate representation: ['select nuclear\_power\_plants.Id', 'select where nuclear\_power\_plants.ConstructionStartAt="Fall" and nuclear\_power\_plants.LastUpdatedAt=2009', 'select where nuclear\_power\_plants.ConstructionStartAt="Spring" and nuclear\_power\_plants.LastUpdatedAt=2010'] \\
\colorbox{yellow}{A:} Lets think step by step. To get the SQL using the intermediate representations, we combine them to form: \\
\colorbox{yellow}{SQL:} [ \textcolor{blue}{SELECT} Id \textcolor{blue}{FROM} nuclear\_power\_plants \textcolor{blue}{WHERE} ConstructionStartAt = 'Fall' AND LastUpdatedAt = 2009 UNION \textcolor{blue}{SELECT} Id \textcolor{blue}{FROM} nuclear\_power\_plants \textcolor{blue}{WHERE} ConstructionStartAt = 'Spring' AND LastUpdatedAt = 2010" ] \\
\# \\
Q: Find the countries and average capacities of all nuclear power plants whose average capacity is greater than 42000. \\
sub-questions:['Find the countries and average capacities of all nuclear power plants, whose average capacity is greater than 42000.'] \\
Intermediate representation: ['select nuclear\_power\_plants.Country, avg(isntructor.Capacity) group by nuclear\_power\_plants.Country', 'select where avg(nuclear\_power\_plants.Capacity) > 42000'] \\
\colorbox{yellow}{A:} Lets think step by step. To get the SQL using the intermediate representations, we combine them to form: \\
\colorbox{yellow}{SQL:} [ \textcolor{blue}{SELECT} Country, AVG (Capacity) \textcolor{blue}{FROM} nuclear\_power\_plants \textcolor{blue}{GROUP BY} Country HAVING AVG (Capacity) > 42000 ] \\
\# \\
Q: display the Name of the nuclear power plant and the difference between its capacity and the year it was constructed for those plants which capacity is within the range 12000 to 18000. \\
sub-questions:['display the Name of the nuclear power plant and the difference between its capacity and the year it was constructed for those plants', 'which capacity is within the range 12000 to 18000.'] \\
Intermediate representation: ['select nuclear\_power\_plants.Name, nuclear\_power\_plants.Capacity - nuclear\_power\_plants.ConstructionStartAt', 'select where nuclear\_power\_plants.Capacity between 12000 and 18000'] \\
\colorbox{yellow}{A:} Lets think step by step. To get the SQL using the intermediate representations, we combine them to form: \\
\colorbox{yellow}{SQL:} [ \textcolor{blue}{SELECT} Name, Capacity - ConstructionStartAt \textcolor{blue}{FROM} nuclear\_power\_plants \textcolor{blue}{WHERE} Capacity BETWEEN 12000 AND 18000 ] \\
\# \\
Q: display the ID for each nuclear power plant and the date on which it stopped operating. \\
sub-questions:['display the ID for each nuclear power plant and the date', 'on which it stopped operating.'] \\
Intermediate representation: ['select nuclear\_power\_plants.Id, max(nuclear\_power\_plants.OperationalTo) group by nuclear\_power\_plants.Id', 'select extra max (nuclear\_power\_plants.OperationalTo)'] \\
\colorbox{yellow}{A:} Lets think step by step. To get the SQL using the intermediate representations, we combine them to form: \\
\colorbox{yellow}{SQL:} [ \textcolor{blue}{SELECT} Id, MAX(OperationalTo) \textcolor{blue}{FROM} nuclear\_power\_plants \textcolor{blue}{GROUP BY} Id ] \\
\# \\
Q: return the smallest capacity for each nuclear power plant. \\
sub-questions:['return the smallest capacity for each nuclear power plant.'] \\
Intermediate representation: ['select min(nuclear\_power\_plants.Capacity), nuclear\_power\_plants.Id group by nuclear\_power\_plants.Id'] \\
\colorbox{yellow}{A:} Lets think step by step. To get the SQL using the intermediate representations, we combine them to form: \\
\colorbox{yellow}{SQL:} [ \textcolor{blue}{SELECT} MIN(Capacity), Id \textcolor{blue}{FROM} nuclear\_power\_plants \textcolor{blue}{GROUP BY} Id ] \\
\# \\
Q: display the id and the total capacity for those nuclear power plants which have at least two reactors. \\
sub-questions:['display the id and the total capacity for those nuclear power plants', 'which have at least two reactors.'] \\
Intermediate representation: ['select nuclear\_power\_plants.Id, sum(nuclear\_power\_plants.Capacity)', 'select where count (nuclear\_power\_plants.*)>=2 group by nuclear\_power\_plants.Id'] \\
\colorbox{yellow}{A:} Lets think step by step. To get the SQL using the intermediate representations, we combine them to form: \\
\colorbox{yellow}{SQL:} [ \textcolor{blue}{SELECT} Id, SUM(Capacity) \textcolor{blue}{FROM} nuclear\_power\_plants \textcolor{blue}{GROUP BY} Id HAVING count(*) >= 2 ] \\
\# \\
Q: Count how many times the capacity of a nuclear power plant is equal to the sum of its status and reactor type. \\
sub-questions:['Count how many times the capacity of a nuclear power plant', 'is equal to the sum of its status', 'and reactor type.'] \\
Intermediate representation: ['select count(nuclear\_power\_plants.*)', 'select where nuclear\_power\_plants.Capacity=nuclear\_power\_plants.Status + nuclear\_power\_plants.ReactorType','select extra nuclear\_power\_plants.Capacity'] \\
\colorbox{yellow}{A:} Lets think step by step. To get the SQL using the intermediate representations, we combine them to form: \\
\colorbox{yellow}{SQL:} [ \textcolor{blue}{SELECT} COUNT(*) \textcolor{blue}{FROM} nuclear\_power\_plants AS T1 JOIN nuclear\_power\_plants AS T2 ON T1.Id = T2.Id \textcolor{blue}{WHERE} T2.Capacity = T1.Status + T1.ReactorType ] \\
\# \\
Q: Find the names of the nuclear power plants with capacity smaller than 3 or capacity greater than 5. \\
sub-questions:['Find the names of the nuclear power plants with capacity', 'smaller than 3 or', 'capacity greater than 5.'] \\
Intermediate representation: ['select nuclear\_power\_plants.Name, 'select where nuclear\_power\_plants.Capacity < 3','select where nuclear\_power\_plants.Capacity > 5'] \\
\colorbox{yellow}{A:} Lets think step by step. To get the SQL using the intermediate representations, we combine them to form: \\
\colorbox{yellow}{SQL:} [ \textcolor{blue}{SELECT} Name \textcolor{blue}{FROM} nuclear\_power\_plants \textcolor{blue}{WHERE} Capacity < 3 OR Capacity > 5 ] \\
\# \\
\colorbox{cyan}{Q: Which country has the most capacities of nuclear power plants?} \\
\colorbox{cyan}{sub-questions:['Which country has the most capacities of nuclear power plants?']} \\
\colorbox{cyan}{Intermediate representation: ['select nuclear\_power\_plants.Country, sum(nuclear\_power\_plants.Capacity) group} \\
\colorbox{cyan}{by nuclear\_power\_plants.Country', 'select order by sum(nuclear\_power\_plants.Capacity) desc limit 1']} \\
\colorbox{cyan}{A: } \\

\end{document}